\documentclass[sigchi]{acmart}

\usepackage{tabularx}
\usepackage{tikz}
\usetikzlibrary{arrows.meta, calc, shapes.geometric}

\usepackage{enumitem}
\setlist[itemize]{leftmargin=*, topsep=0pt}
\usepackage{subcaption}
\usepackage{longtable,booktabs,array}

\AtBeginDocument{%
  }

\setcopyright{acmlicensed}
\copyrightyear{2026}
\acmYear{2026}
\begin{document}

\title{Palimpsestic Accountability in the US Federal AI Inventories}

\author{Dipto Das}
\affiliation{%
  \department{College of Computing and Information Science}
  \institution{Cornell University}
  \city{Ithaca}
  \state{New York}
  \country{United States}
}
\email{dd749@cornell.edu}

\author{Syed Ishtiaque Ahmed}
\affiliation{%
  \department{Department of Computer Science}
  \institution{University of Toronto}
  \city{Toronto}
  \state{Ontario}
  \country{Canada}
}
\email{ishtiaque@cs.toronto.edu}

\author{Aditya Vashistha}
\affiliation{
  \department{College of Computing and Information Science}
  \institution{Cornell University}
  \city{Ithaca}
  \state{New York}
  \country{United States}
}
\email{adityav@cornell.edu}

\author{Shion Guha}
\affiliation{%
  \department{Faculty of Information}
  \institution{University of Toronto}
  \city{Toronto}
  \state{Ontario}
  \country{Canada}
}
\email{shion.guha@utoronto.ca}




\begin{abstract}
  An AI inventory is a structured, often public record in which a government or organization lists and describes the AI systems it uses or plans to use. Studies of single AI inventory releases describe what systems do but say little about how they are governed over time. We linked entries across three releases of the US Federal AI Use Case Inventory (2023--2025), compared answers only where both releases required them, and analyzed governance accounts through a dual deductive-inductive approach, guided by the Algorithmic Decision-Making Adapted for the Public Sector (ADMAPS) framework. As the inventory grew from 710 to 3,611 entries from 2023 to 2025, fewer required answers were left blank, but accountability was rewritten through three mechanisms. Each release \emph{redrew} the scope of the questions; agency-assigned classifications \emph{gated} what was reported; and answers were also \emph{overwritten}, with gains and losses concentrated in single agencies and little account of human oversight and judgment. These findings show that each release redraws the boundary of accountability and leaves the public to \emph{re-stitch} the records, a condition we term \textbf{palimpsestic accountability}. We recommend versioned records, reasons for every reclassification, public files that distinguish kinds of missing answers, and oversight documented as conditions to improve public sector AI accountability.

\end{abstract}

\begin{teaserfigure}
  \centering
  \includegraphics[width=\textwidth]{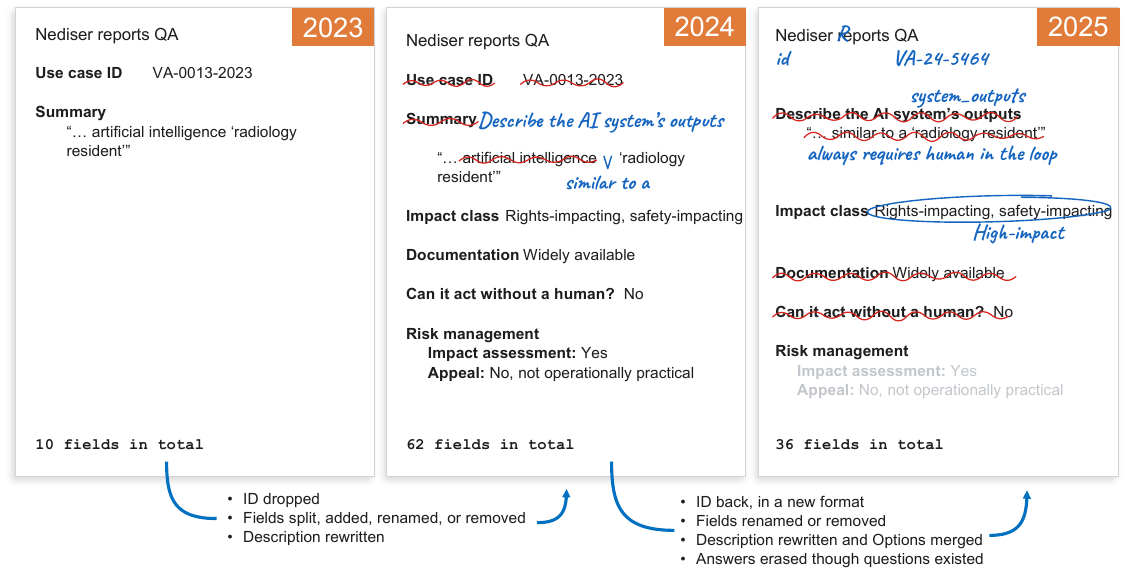}
  \caption{A palimpsest is a manuscript scraped and written over, with traces of earlier text still visible. One AI use case reported across three releases (2023--2025) of the US Federal AI Use Case Inventory, where each release writes over the last. Annotations show what changed in the entry from the previous release.}
  \label{fig:teaser}
\end{teaserfigure}

\maketitle

\section{Introduction}
Governments, companies, and international organizations have published hundreds of guidelines committing artificial intelligence (AI) to principles like fairness, accountability, and transparency (FAccT)~\cite{jobin2019global, correa2023worldwide}. While FAccT research has advanced documentation practices (e.g., datasheets and model cards~\cite{gebru2021datasheets, mitchell2019model}), AI inventory (or register) is a key way governments extend these practices in the public sector. An AI use case inventory is a structured, often public record in which a government or organization lists its current and planned AI use cases and describes each one's purpose, technical properties, context, and governance~\cite{gpai2024algorithmic, nieuwenhuizen2024algorithm}. In the United States (US), a 2020 census found 157 AI use cases across 64 of the 142 federal agencies~\cite{engstrom2020government}. An Executive Order and the Advancing American AI Act subsequently required agencies to inventory their AI use cases~\cite{president2020promoting, peters2021advancing}. When the Office of Management and Budget (OMB) released the consolidated 2025 inventory, news coverage reported that the number of federal AI uses had more than doubled in a year~\cite{aldane2026number, kelley2026agencies}. Under current OMB guidance, each inventory entry is an agency's answers to a questionnaire about one AI use, including its purpose, outputs, lifecycle stage, impact classification, and risk-management practices. Some of these questions were required for every entry, whereas others were only for certain ones (e.g., whether a use case had an impact assessment or appeal process).

Prior research has mostly examined public AI registers one release at a time~\cite{nieuwenhuizen2024algorithm, das2026global, das2026bureaucratic, pi2026understanding}. Such studies can show what a register discloses or omits, but not whether its transparency and silence persist, deepen, or move between releases. In the US context, research has mostly raised concerns about unpublished inventories, inaccurate data, and blank fields~\cite{gao2023artificial, anexries2026oneyear}. Others have used the reported cases in the inventories to gauge Federal AI adoption~\cite{wirtschafter2025assessing} or compared different AI policies~\cite{lurie2026triangulating, alanoca2025comparing}. However, after three releases of the US AI inventory, we know comparatively little about how algorithmic decision-making fits into federal agency bureaucracies and the people who carry it out, or how agencies' accounts of the same AI uses change over time.

In this paper, we conducted a longitudinal mixed-methods study of the US Federal AI Use Case Inventory across three reporting cycles (2023--2025), examining what it tells the public about algorithmic decision-making in federal agencies and how agencies classified, justified, and described the same AI as reporting requirements changed. We linked entries describing the same systems across years, and combined descriptive statistics with a dual deductive-inductive qualitative analysis, conceptually sensitized by the Algorithmic Decision Making Adapted for the Public Sector (ADMAPS) framework~\cite{saxena2021framework}. Beyond an overview of all entries, we focused on the uses federal policy prioritized for scrutiny and required further governance-related information and justification.
 
We found that each release \emph{redrew} the scopes of questions that agencies were asked, and classification rules \emph{gated} what information was reported. About one in six (44 of 272) AI uses classified as ``rights-impacting" or ``safety-impacting" in 2024 fell outside the ``High Impact" (HI) category in 2025,  which exempted them from reporting impact assessments, monitoring, and appeals, and most gave no reason or a template sentence. Answers were also \emph{overwritten}. For four questions required of the same AI use cases in both 2024 and 2025, one agency contributed 89\% of the answers that newly appeared in 2025, and another accounted for 84\% of those that disappeared, so whether an entry gained or lost information depended more on which agency reported it than on the AI system it described. Entries often named who could review or correct an output, but rarely explained how human officers weighed that output against other evidence, and fewer than one in five high-impact uses reported established operator training. We describe this process through the metaphor of a \textbf{palimpsest}, a manuscript page scraped and written over (as in Figure~\ref{fig:teaser}), on which traces of the earlier writing remain partly legible~\cite{dillon2007palimpsest}. Based on our findings, we propose the term \textbf{\emph{palimpsestic accountability}} to describe how the accountability an inventory supports depends not only on what each release discloses, but also on whether its successive accounts remain identifiable, traceable, and connected. Overall, this paper makes three contributions:

\begin{itemize}
    \item First, we provide an \textbf{empirical} account of the US Federal AI Use Case Inventory, answering \emph{what it discloses about algorithmic decision-making in the public sector} and \emph{how its accounts changed across releases}. Because the inventories lack persistent identifiers, we linked the same AI uses across releases by combining automated matching with manual review, and publish the \emph{linked dataset} (available upon the paper's acceptance) to support future longitudinal analyses.
    \item Second, we propose the \textbf{concept} of \emph{``palimpsestic accountability"} to describe how successive releases shift the work of reconnecting accounts from agencies to the public.
    \item Third, we offer \textbf{recommendations} for public AI inventories, such as persistent identifiers and documented revisions, explanations whenever a use is reclassified, guides that distinguish kinds of missing answers, and accounts of conditions supporting human oversight.
\end{itemize}%\clearpage
\section{Literature Review}
In this section, we first discuss how various documentation artifacts materialize algorithmic accountability. Then, we focus on AI inventory, a type of accountability artifact, and discuss how it has been used and examined across contexts, ranging from global to national to municipal governments. %, and why the longitudinal study presented in this paper is warranted.

\subsection{Documentation as Algorithmic Accountability Artifacts: The Case of AI Inventory}
Documentation artifacts (e.g., datasheets~\cite{gebru2021datasheets}, model cards~\cite{mitchell2019model}) are one of the main ways to make algorithmic accountability concrete. Public sector institutions scale this documentation logic from a single model or dataset to an institution or a government's entire portfolio of algorithmic systems, and publish various algorithmic accountability artifacts, such as AI registers/inventories~\cite{sengupta2022algorithm}, internal audit reports~\cite{raji2020closing}, and impact assessments~\cite{watkins2021governing}. In this paper, we focus on AI inventories, which are standardized, often public lists in which governments describe the AI systems they use, usually including each system's name, purpose, responsible units, data, and technical capacities, among other information (see Table~\ref{tab:fields-by-year} in Appendix~\ref{app:fields} for example fields the US reported). 

% standardize information about provenance, intended use, limitations, and responsible actors so that it can circulate among developers, organizations, regulators, and publics. 

However, making a system's properties visible does not by itself make it understandable~\cite{ananny2018seeing}, and more information does not necessarily increase trust~\cite{kizilcec2016much}. Disclosures are situated communications whose content, format, and intention shape what audiences can know~\cite{eyert2023rethinking, norval2022disclosure}. The sections on limitations and evaluation being the least completed in these documents~\cite{liang2024systematic} underscores this concern. Such reporting gaps often stem from weak incentives, scarce resources, a lack of good-faith participation, and poor integration with organizational workflows~\cite{winecoff2025improving, selbst2021institutional}.

Moreover, documentation does not inherently make the systems accountable to the public. For example, these documentation artifacts often record metrics (e.g., ``impact"), which sometimes become an organizationally legible metric without representing how people experience and evaluate this construct~\cite{metcalf2021algorithmic, das2026asymmetries}. Claims of scale or completeness can hide the complex coordination of humans, organizations, and technologies behind these artifacts~\cite{lee2006human}, along with questions like how its data are produced and whose labor they depend on~\cite{denton2021genealogy, scheuerman2025data}. Reading datasets and disclosure datasets critically reveals the classificatory and political choices they encode~\cite{poirier2021reading, poirier2022accountable}. For example, the Digital Services Act (DSA) Transparency Database, one of the European Union (EU)'s major accountability artifacts, allows discretion over what to disclose~\cite{kaushal2024automated} to coordinate reporting among entities with varied organizational practices. Such flexibility can create inconsistencies across reporting periods~\cite{groesch2025big, shahi2025year}, and often goes unreconciled with harmonized templates~\cite{trujillo2026disarranged}. Moreover, the underlying classificatory logics may also change over time~\cite{bowker2000sorting}. For example, in the US context, the ``rights-impacting" and ``safety-impacting" AI were consolidated into ``high-impact" AI~\cite{friedler2025omb}, which raises questions about how agencies applied the new category to particular uses. Scholars emphasize that classifications never sort every case cleanly~\cite{bowker2000sorting}, and hard-to-classify AI systems sit at the edges~\cite{golpayegani2023high}, which this paper dubs ``classification boundary". Their risks, impacts, outcomes, etc. may be reported with different levels of detail depending on their quantifiability~\cite{kaminski2023regulating}. Recent work brings the temporal concerns to AI documentation: deprecated datasets often continue to circulate without standardized identification and communication~\cite{luccioni2022framework}, and deployed models can change without corresponding updates~\cite{abraham2026silent}.

% However,  .

% However, studies of public AI registers have rarely examined how their records, and the categories that structure them, change across reporting cycles.

\subsection{Public AI Registers around the World and the US AI Use Case Inventory}
Since Amsterdam and Helsinki launched some of the first municipal registers in 2020~\cite{haataja2020public, floridi2020artificial}, similar initiatives have emerged in French cities~\cite{ogp2021building}, Chile~\cite{lapostol2023algorithmic}, the UK~\cite{zilka2023exploring, pi2026understanding}, Canada~\cite{gc2025register}, and other municipal and national governments~\cite{sieber2026building}. A 2024 report for the Global Partnership on AI identified 69 active public algorithm repositories worldwide~\cite{gpai2024algorithmic}. While registers are presented as a way to establish AI as a transparent and trustworthy public service~\cite{floridi2020artificial}, scholars have warned that they can decontextualize and depoliticize systems whose harms in the digital welfare state were already documented~\cite{cath2022dutch}. Empirical studies found selective disclosure, limited access, and immature, highly variable register-management practices in Dutch organizations~\cite{nieuwenhuizen2024algorithm, schuitemaker2024realizing}, and surfaced issues, such as wrongful benefit denials, that a register entry itself did not reveal~\cite{detroya2026coconstructing}. A comparative study of Chinese and British AI registers showed that registers can serve functions ranging from public disclosure to pre-market approval and regulatory coordination, depending on their design~\cite{pi2026understanding}. Across 72 countries, registers share a descriptive core but rarely request information about appeals, risks, or external evaluation~\cite{das2026global}. Most of these studies analyzed one register at a time. Peljto et al.'s finding that two German platforms published no update obligations and did not version their entries~\cite{peljto2026are} suggests why change across releases has been difficult to study.

Research on the US Federal AI Use Case Inventory has focused mainly on implementation and completeness, finding that roughly half or more of federal agencies had not published the required inventories~\cite{lawrence2023bureaucratic} and that 15 of the 20 inventories the Government Accountability Office reviewed contained incomplete or inaccurate data~\cite{gao2023artificial}. Nearly half of agencies' compliance plans for OMB's 2024 memorandum did not explain how their inventories would be comprehensive~\cite{wang2025assessing}, and civil society reviews have documented inconsistent and blank risk designations~\cite{anexries2025exploring, anexries2026oneyear}. Others have used the inventory to gauge federal adoption itself~\cite{wirtschafter2025assessing}. While these studies assessed each inventory release as a snapshot, they leave an open question about what happens to an AI use's accounts when the same agency reports it under revised questionnaires in different releases. Lurie et al.'s work is the closest precedent addressing this temporal question. They triangulated the AI Use Case Inventory with two other ``federal AI transparency regimes"~\cite{lurie2026triangulating}, namely system-of-records notices and information collection requests. They matched 2024 entries to 2025 entries. They highlighted that the absence of persistent identifiers, uneven granularity, and the annual reporting cycle make it difficult to track systems across regimes and over time. Whereas their object was disclosure fragmented across three regimes, we look into a single regime, i.e., AI Inventory, fragmenting against itself over time, by re-describing the same use cases from one release to the next.

A sociotechnical reading of these accounts can surface how a system and its description are arranged with people, rules, and organizational routines, as highlighted by prior work in the public sector contexts like social benefits~\cite{eubanks2018automating, redden2022automating}, child welfare~\cite{kawakami2022improving, saxena2023rethinking}, homelessness services~\cite{moon2025datafication}, policing~\cite{selten2023just}, higher education~\cite{mcconvey2024not}, etc. Hence, we draw on the ADMAPS framework~\cite{saxena2021framework}, synthesized from public administration, science and technology studies, and human-computer interaction scholarship, to examine these accounts. Instead of treating public sector algorithms as stand-alone technical artifacts, it examines the intersection of three dimensions: \emph{bureaucratic processes} (resources and constraints, administration and training, and laws and policies), \emph{algorithmic decision-making} (relevant data, types of decision support, and the degree of uncertainty), and \emph{human discretion} (professional expertise, value judgments, and heuristic decision-making). Das and colleagues used it to show that Canada's federal AI Register foregrounds technical capabilities and efficiency while leaving human discretion, training, and uncertainty management largely unaccounted for~\cite{das2026bureaucratic}. We transfer and extend that single-release analysis to the American context, attending to how changes in reporting rules, descriptions of algorithmic roles, and representations of human judgment together shape what the public can examine over time.

% Under \emph{bureaucratic processes}, we examined reporting obligations, institutional support, and governance procedures in relation to \emph{laws and policies}, \emph{resources and constraints}, and \emph{administration and training}. Under \emph{algorithmic decision-making}, we considered inputs, outputs, and stated limitations through \emph{relevant data}, \emph{types of decision-support}, and \emph{degree of uncertainty}. Under \emph{human discretion}, we examined qualifications, priorities, contextual reasoning, and intervention in relation to \emph{professional expertise}, \emph{value judgments}, and \emph{heuristic decision-making}.%\clearpage
\section{Methods}
To trace how the US AI use case inventories described the same systems across its 2023, 2024, and 2025 releases, we first collected data and reconstructed comparable questions and records (Section \ref{sec:data-collection}), and then measured the answers quantitatively and read selected entries closely (Section~\ref{sec:data-analysis}).

\subsection{Data Collection and Preprocessing}\label{sec:data-collection}

\subsubsection{Data Sources and Research Scope}
We used the official inventories maintained by OMB. The 2023 and 2024 consolidated files came from OMB's 2024 inventory repository\footnote{\url{https://github.com/ombegov/2024-Federal-AI-Use-Case-Inventory/}}, and the 2025 files came from its 2025 repository\footnote{\url{https://github.com/ombegov/2025-Federal-Agency-AI-Use-Case-Inventory/}}, as of April 2026. Each row of a release, dubbed an \emph{entry}, describes one \emph{AI use case}. The files contained 710 entries with 10 fields in 2023, 2,133 entries with 62 fields in 2024, and 3,611 entries with 36 fields in 2025 (Appendix~\ref{app:results}, Fig.~\ref{fig:agencies}). The 2025 release also included a separate inventory of commercial-off-the-shelf (COTS) AI, recording whether each agency used those for some of the common tasks. We included this file in our comparison of reporting arrangements, but excluded it from entry-based analysis because it describes agencies' practices in aggregate rather than by individual use case.

\subsubsection{Comparing Questions and Requirements across Years}\label{sec:comparing_req}
For each year, we examined the annual submission instructions, data dictionaries\footnote{Definitions of each field, its permitted answers, and when it was required}, released schemas, and policy documents referenced in those materials. Reading them together allowed us to distinguish what agencies were asked to submit from what appeared in the consolidated public files. Because each release asked different questions, we first mapped each question to its counterparts in other years (Table~\ref{tab:fields-by-year} in Appendix~\ref{app:fields}). We found that the questions across years changed in several ways. While some new questions were introduced, some existing ones were removed, renamed, split, merged\footnote{A 2024 use was rights- or safety-impacting under OMB M-24-10 if its output served as a principal basis for decisions or actions significantly affecting people's rights or access to critical services, or affecting human safety, the environment, critical infrastructure, or strategic assets; M-25-21 merged the two into a single HI category defined by the same kinds of effects~\cite{friedler2025omb}.}, or moved to a separate file (Table~\ref{tab:question-changes} in Appendix~\ref{app:fields}).

% Questions changed in several ways, such as: (a) \emph{Introduced}: the 2025 inventory asked whether operators of high-impact (HI) uses had been trained; (b) \emph{Removed}: the 2024 question about whether a use could act without human involvement had no 2025 counterpart; (c) \emph{Renamed}: the field recording a use's \texttt{Bureau} in 2024 became \texttt{agency\_bureau} in 2025; (d) \emph{Split}: the single 2023 summary became separate questions about purpose and benefits and about outputs in 2024, and the 2025 inventory split purpose and benefits into the problem addressed and the expected benefits; (e) \emph{Merged}: the 2024 inventory separately asked whether a use was rights- and safety-impacting, whereas the 2025 inventory replaced the two categories with a single HI category;\footnote{A 2024 use was rights- or safety-impacting under OMB M-24-10 if its output served as a principal basis for decisions or actions significantly affecting people's rights or access to critical services, or affecting human safety, the environment, critical infrastructure, or strategic assets; M-25-21 defined the 2025 HI category by the same kinds of effects~\cite{friedler2025omb}.} (f) \emph{Moved}: the 2024 inventory asked whether each use was a common commercial AI product, whereas the 2025 release reported such products in the separate COTS inventory.

A use case's lifecycle stage and impact classification often decided which questions that entry had to answer. Per OMB guidance, the impact classification determined whether a use case had to answer governance questions, such as impact assessment and appeals. For example, in the 2025 inventory, a use case labeled ``Presumed High-Impact, but Not High-Impact" (PHIN) had to provide a written justification of why it did not meet the HI definition despite being deemed rights- or safety-impacting in the previous year, whereas one labeled not-HI did not. For each entry (i.e., a use case) and question, we then determined whether that year's rules required an answer. For each entry, we labeled each question as \emph{required}, \emph{not required}, \emph{optional}, or \emph{unspecified} (see examples in Table~\ref{tab:requirement-examples} in Appendix~\ref{app:requirement_check}). The 2023 release did not include a data dictionary, so we could not determine which questions an agency had to answer, and compared requirements only between 2024 and 2025.

In addition, we also labeled every answer into one of six states: (i) \emph{Present} (any substantive answer), (ii) \emph{No} (an explicit negative; e.g., identifying a use case as ``not HI"), (iii) \emph{Not applicable} (e.g., some questions not being applicable for use cases that are not HI), (iv) \emph{Blank} (e.g., empty cells, \texttt{null}), (v) \emph{Unknown} (e.g., an agency explicitly describing the answer to be determined), or (vi) \emph{Withheld} (e.g., an answer being redacted for legal or cybersecurity reasons) (detail in App~\ref{app:response-classification}).

\subsubsection{Matching the Same Uses across Releases}
To analyze how an AI use case changed over the years, we needed a mechanism to track it. However, the 2024 file had no ID column. Though the 2023 and 2025 files had an ID column, their formats differed across years and agencies: \texttt{"<agency>-NNNN-2023"} in 2023 and mostly \texttt{"<agency>-N"} in 2025, while 704 entries that year had none. Moreover, AI use cases were named differently across releases (see Figure~\ref{fig:teaser}). Because each inventory contains many entries, manual matching was not possible either. We therefore adopted a mixed-method approach. First, we linked entries whose identifiers or titles matched exactly, applying four rules in order: (1) identical identifiers; (2) identical titles; (3) identical titles with different bureaus\footnote{E.g., two Department of Justice (DOJ) use cases were both titled ``Thomson Reuters CLEAR", one at the Bureau of Alcohol, Tobacco, Firearms and Explosives and another at the US National Central Bureau, respectively.}; and (4) titles in which one began with the other\footnote{E.g., The same system was named ``AI based Semantic Search for National Archives Catalog" in 2024 and ``AI based Semantic Search for National Archives Catalog (aka ArchiAI)" in 2025}. Second, for the remaining entries, we scored every possible pair on the similarity of their titles, descriptions, and bureaus, and used the Hungarian algorithm to propose the best match for each entry~\cite{crouse2016implementing, jaro1989advances}. Third, the first author reviewed the proposed pairs with high similarity scores and marked each as accept, reject, or uncertain (see examples in Table~\ref{tab:review-examples}). Appendix~\ref{app:methods-supplement} and Table~\ref{tab:methods-linkage-review} provide details on text normalization, score weights, thresholds, and counts. Together, the exact and accepted matches yielded 254 matched pairs between 2023 and 2024, 1,332 between 2024 and 2025, 227 between 2023 and 2025, and 150 across all three years. We did not assume entries without an accepted match were new or discontinued uses, since a use could also go unlinked if it was renamed, merged, split, or moved between bureaus beyond what our rules for matching could detect.

% \begin{table}[!ht]
%     \centering
%     \footnotesize
%     \caption{An example of changing IDs and titles for the same AI use case in different releases.}
%     \label{tab:example_of_change}
%     \begin{tabular}{p{.07\linewidth}p{.33\linewidth}p{.25\linewidth}p{.25\linewidth}}
%     \toprule
%          & \textbf{2023} & \textbf{2024} & \textbf{2025} \\
%     \midrule
%         \textbf{ID} & NARA-0003-2023 & (no ID column) & NARA - 0006 \\\hline
%         \textbf{Title} & Semantic Search for National Archives Catalog - an Artificial Intelligence (AI) / Machine Learning (ML) Pilot Program & AI based Semantic Search for National Archives Catalog & AI based Semantic Search for National Archives Catalog (aka ArchiAI)\\
%     \bottomrule
%     \end{tabular}
% \end{table}

\begin{table*}[!ht]
  \caption{Examples of review decisions on proposed matches. Scores are similarity scores between 0 and 1.}
  \label{tab:review-examples}
  \footnotesize
  \begin{tabular}{lp{.3\linewidth}p{.25\linewidth}p{.14\linewidth}p{.12\linewidth}}
    \toprule
    Years & Earlier title & Later title & (Overall, Title, Desc.) similarity & Human review \\
    \midrule
    2024--25 & Taxpayer 90 (Palantir prototype) & Taxpayer 90 & (0.51, 0.62, 0.00) & Accept\\
    2024--25 & Climate Change Assessment Tool & DHS Asset Assessment Tool & (0.48, 0.51, 0.18) & Reject\\
    2023--24 & NIH Grants Virtual Assistant & NBS Virtual Assistant & (0.56, 0.64, 0.08) & Uncertain\\
    \bottomrule
  \end{tabular}
\end{table*}

\subsubsection{The Linked Dataset}
These steps produced the linked dataset on which all later analyses relied. It contains (1) the 6,454 entries, each with its agency, lifecycle stage, and impact classification as reported and as grouped for comparison across years, plus the requirement label and response state of every question; and (2) the accepted matches, each with the rule or similarity scores that proposed it and, for fuzzy matches, the review decision.

% These steps produced the linked dataset on which all later analyses relied. It has two parts: (1) the 6,454 entries of the three releases, each with its harmonized fields, a requirement label for each question, and a response state for each answer; and (2) a table of matched pairs, recording for each pair the two entries, the rule or score that proposed the match, and the review decision. We built the dataset and computed all statistics and figures in Python.

\subsection{Data Analysis}\label{sec:data-analysis}
\subsubsection{Quantitative Analysis}
First, our quantitative analyses describe what agencies reported about their AI uses across the annual inventory releases, using visualizations and descriptive statistics.

Second, we examined how the lifecycle stages and impact classifications of the same uses changed over time. A use's \emph{lifecycle stage} is where the agency reported it in its development, from planning to retirement. We grouped the 2024 and 2025 stages into four common stages: \emph{pre-deployment} (2024's Initiated and Acquisition and/or Development, 2025's Pre-deployment); \emph{pilot} (2024's Implementation and Assessment, 2025's Pilot); \emph{deployed} (2024's Operation and Maintenance, 2025's Deployed); and \emph{retired} (Retired in both years). The 2023 inventory offered no fixed options, so agencies described stages in their own words. A use's \emph{impact classification} is the agency's determination of whether the use's outputs could significantly affect people's rights or safety, which decided whether it had to answer the governance questions. The 2023 inventory did not report impact classifications. The options were rights-impacting, safety-impacting, both, or neither in 2024, and HI, PHIN, or not HI in 2025. We cross-tabulated each matched use's stage and impact classifications in 2024 versus 2025 (Table~\ref{tab:results-lifecycle} and Figure~\ref{fig:classification_transition} in Appendix~\ref{app:results}).

Third, we examined how agencies answered the same questions from one year to the next. We followed columns that recurred between 2024 and 2025. For each use found in both years, we recorded whether an answer to each column was required in both years, in only one of them, or in neither. We compared answers only among uses that had to answer in both years, because a blank answer to a question a use no longer had to answer reflects a change in the rules rather than an undisclosed answer. For example, of the 1,332 matched uses, 548 had to report their procurement method in both years, but only 26 had to report on impact assessment.

\subsubsection{Qualitative Analysis}
The counts show whether an answer remained available, but not what it said or how an agency's account of a use changed. We therefore read two sets of entries closely, using the ADMAPS framework~\cite{saxena2021framework} to organize what they said about the organization, the algorithm, and the people involved. The first set consisted of the 171 matched uses that were deployed in both 2024 and 2025 and classified as rights- and/or safety-impacting in 2024. The second set included all 110 entries in the 2025 PHIN category, 109 of which provided a justification. The two sets overlap in 13 entries. We chose these sets for further qualitative analyses because they hold the inventory's most detailed accounts of governance. The first set had to answer governance questions such as whether the use had an impact assessment, independent review, and ongoing monitoring; whether it could operate without human involvement; and whether people were notified and could appeal or opt out. The second set required justifications of re-classifications (e.g., from rights/safety-impacting, both, or none to HI, PHIN, or not HI, as previously described in Sec.~\ref{sec:comparing_req}). Because these two sets cover only the use cases with the most governance obligations, we also read these entries' answers to other fields, such as data descriptions and benefits.

We used a dual deductive-inductive approach, commonly used in research at the intersection of computing and human, organizational, and social factors (e.g.,~\cite{goldkuhl2024aspects, kiene2026relational, nolte2020support}). ADMAPS served as ``sensitizing concepts"~\cite{bowen2006grounded}: it directed our reading to how each entry described the organization around the system (\emph{bureaucratic processes}, such as resources, training, and policies), the system's role (\emph{algorithmic decision-making}, such as its data, outputs, and limitations), and the people who used its outputs (\emph{human discretion}, such as who reviewed outputs, with what expertise, and on what basis). These dimensions conceptually overlap with many of the inventories' questions. The first author inductively coded each entry and matched-pair, refining the codes through discussions with the other three authors and recording disagreements in analytic memos until resolution. We then grouped these codes into the three mechanisms that organize our results. Following guidance for collaborative interpretive analysis~\cite{mcdonald2019reliability}, we did not calculate inter-coder reliability.

\subsection{Limitations}
This study has several limitations. First, our analysis draws on agencies' public accounts of their AI uses, not direct observation of their practices. Second, our matches depend on scoring rules and thresholds that we set. Because no benchmark exists for matching entries across releases of public-sector AI inventories, we adapted general record-linkage methods, as~\cite{lurie2026triangulating} did. Manual review of proposed pairs reduced false matches, but use cases that were substantially renamed or changed may have remained unmatched. Third, we deliberately focused our close reading on HI and PHIN uses, where the inventory requires the fullest governance accounts. Lower-impact uses, which make up most entries, may describe their systems and oversight differently, and other frameworks could foreground aspects ADMAPS did not.%\clearpage
\section{Results}
Our analysis found that from 2023 to 2025, the US federal AI use case inventories widened visibility into the range of federal AI uses while narrowing visibility into how any particular use was reclassified, described, or overseen. We organize the results by the three mechanisms through which accountability was rewritten across releases: \emph{redrawing the questions, gating uses by impact, and overwriting earlier answers}. Within each, we report what the releases disclosed about the ADMAPS dimensions and their aspects (e.g., \emph{resources and constraints} within \emph{bureaucratic processes}).

\subsection{Redrawing: Questions Added, Removed, and Reformatted}
As noted earlier, each release changed the number of questions (10 in 2023, 62 in 2024, and 36 in 2025) and how it asked them (see Tables~\ref{tab:fields-by-year} and~\ref{tab:question-changes} in Appendix~\ref{app:fields}). In the last two years, the inventory \emph{added} questions on training and classification justification, \emph{removed} questions on computing resources, documentation, and human involvement, and \emph{reformatted} those on AI techniques, development routes, and COTS AI tools.

\subsubsection{Removal of Questions about Resources, Documentation, and Human Involvement}
\begin{figure*}[t]
  \scriptsize
  \centering
  \begin{minipage}[t]{0.5\linewidth}\vspace{0pt}
    \begin{subfigure}{\linewidth}
        \centering
        \includegraphics[width=\linewidth]{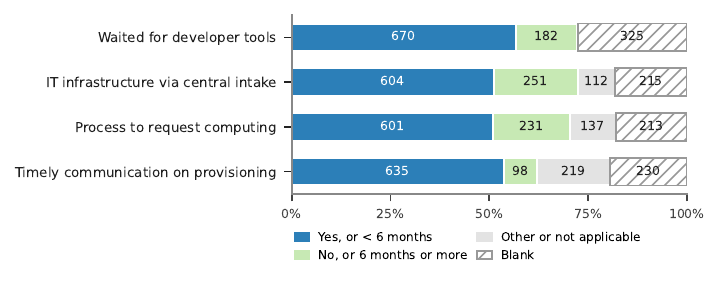}
        \caption{2024 resource questions for AI in development/later.}
        \label{fig:resources-questions}
    \end{subfigure}
  \end{minipage}\hfill
  \begin{minipage}[t]{0.46\linewidth}\vspace{0pt}
    \begin{subfigure}{\linewidth}
        \centering
        \includegraphics[width=\linewidth]{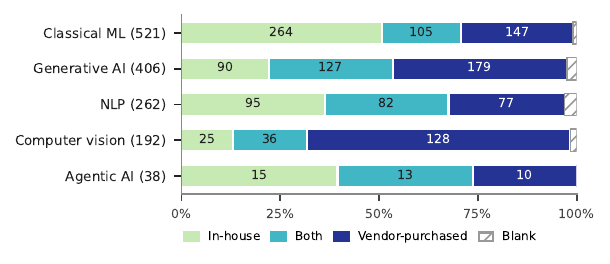}
        \caption{2025 development route by AI techniques.}
        \label{fig:resources-procurement}
    \end{subfigure}
  \end{minipage}
  \caption{Information about AI use cases' development routes from the 2024 and 2025 inventories.}
  \label{fig:development-routes}
\end{figure*}

Many entries, regardless of year, justified AI use by the staff time it would save. However, the 2025 inventory removed the questions about the \emph{resources and constraints}, such as personnel, tools, budgets, or computing capacity that agencies committed, lacked, or redirected to sustain it. Entries described offices ``already operating lean" (DOJ-0353), staff ``overburdened by manual document handling" (USDA-070), and AI adopted because ``with limited staff, we use AI to summarize incoming public complaints" (DOJ-0315). Only the 2024 inventory asked about the organizational resources behind AI uses. Its four questions, asked of use cases in development or later lifecycle stages that did not rely solely on a COTS AI, covered: (i) \emph{how long the use had waited for basic computing and developer tools}; (ii) \emph{whether its information technology (IT) infrastructure was centrally provisioned}; (iii) \emph{whether it had a process to request computing resources\footnote{E.g., cloud computing credits or computing hardware}}; and (iv) \emph{whether communication about provisioning had been timely}. Figure~\ref{fig:resources-questions} shows that about one in five entries reported no centralized intake or no process to request, indicating uneven access to computing resources. DOJ left all four questions blank for its 189 entries. With these questions removed altogether, a 2025 entry can claim that AI relieves a staffing shortage while falling outside the scope of accountability regarding the resources that AI use case itself requires and constraints it imposes.

The 2025 inventory also removed two 2024 questions that asked about a system in use. The first asked whether documentation such as performance metrics and intended use was available for review: only 17.2\% of 2024 entries reported complete documentation, 31.9\% limited documentation, and 16.3\% none. The second asked whether the AI could act without direct human involvement, to which more than 60\% answered ``no". Removing both these questions left the 2025 inventory with no question about how well a system was documented and whether a person stood between its outputs and their effects. Since an AI system cannot be held legally liable, that person becomes the point where responsibility and accountability for that AI use would rest~\cite{bryson2017and}, but becomes illegible from the latest inventory.

\subsubsection{Reformatting of Question Framing and Answer Types, Options, and Arrangements}
Both 2024 and 2025 asked whether a use was developed in-house, by contractors or vendors, or through a combination, with slightly different wording and eligibility. After reformatting, among entries required to answer, blank responses fell from about one in five in 2024 to under 3\% in 2025. That helped us to see how the shares for all three development routes increased: in-house (31.8\% to 34.1\%), contractor (35.3\% to 37.3\%), and combination (11.7\% to 26.1\%) between 2024 and 2025. Among the 425 matched uses that answered in both years, the share with any in-house involvement rose from 60.0\% to 77.4\%. Because the 2025 options were reworded and additional questions about specific AI techniques were asked (more on that in the next paragraph), we could get more granular transparency into the government's in-house technical capability as well as its reliance on and collaboration with vendors (see Figure~\ref{fig:resources-procurement}).

\begin{figure*}[t]
  \centering
  \scriptsize
  \begin{minipage}[t]{0.55\linewidth}\vspace{0pt}
    \begin{subfigure}{\linewidth}
        \centering
        \includegraphics[width=.8\linewidth]{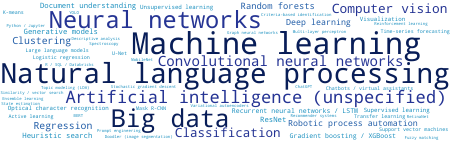}
        \caption{Techniques named in 2023.}
        \label{fig:techniques-wordcloud}
    \end{subfigure}\\[4ex]
    \begin{subfigure}{\linewidth}
      \centering
      \includegraphics[width=\linewidth]{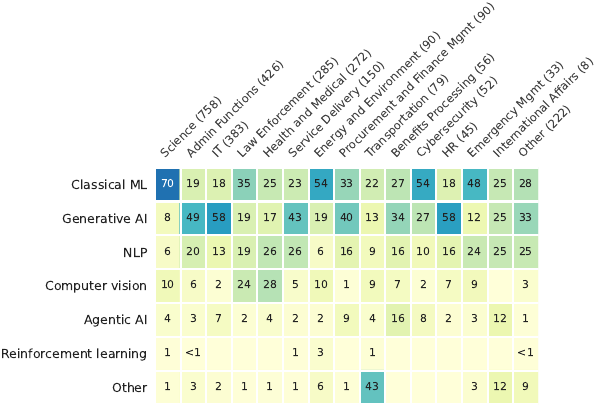}
      \caption{AI techniques' percentages by topic areas in 2025.}
      \label{fig:techniques-topic}
    \end{subfigure}
  \end{minipage}\hfill
  \begin{minipage}[t]{0.43\linewidth}\vspace{0pt}
    \begin{subfigure}{.93\linewidth}
      \centering
      \includegraphics[width=\linewidth]{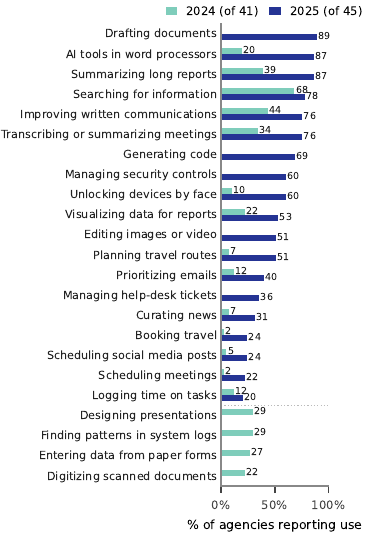}
      \caption{\% agencies using COTS AI in tasks.}
      \label{fig:techniques-cots}
    \end{subfigure}
  \end{minipage}
  \caption{AI techniques and commercial AI tasks reported in the inventories.}
  \label{fig:techniques}
\end{figure*}

In 2023, agencies were asked to name techniques in their own words (see a word cloud in Figure~\ref{fig:techniques-wordcloud}); in 2024, they were not asked this question at all; and in 2025, they were asked to choose techniques from a fixed list, such as classical machine learning (ML), generative AI, natural language processing (NLP), computer vision, and agentic AI. The share of entries reporting a technique rose from 44.4\% in 2023 to 99.3\% in 2025. In 2025, agencies also reported the entries' application domains/topics from a fixed list (e.g., science, IT). AI techniques clustered by these topics (Figure~\ref{fig:techniques-topic}): classical ML accounted for 70.2\% of science entries and more than half of energy and environment (54.4\%) and cybersecurity (53.8\%) entries; generative AI accounted for more than half of IT (58.2\%) and human resources-related (57.8\%) entries; and computer vision was most common in health and medical (27.9\%) and law enforcement (24.2\%) entries. However, the 2025 options, such as ``classical ML'' or computer vision, say less about how a system works than 2023's descriptions, which named methods such as convolutional neural networks or regression.

The commercial AI question moved from individual uses to agencies. In 2024, each use-case entry asked whether the use was one of 19 listed commercial AI tasks (e.g., summarizing a long report, searching for information). In 2025, a separate file asked each agency whether it used commercial AI for each of 20 tasks, without describing any individual use. Among the 14 tasks listed in both years, the share of agencies reporting each task increased (Figure~\ref{fig:techniques-cots}): for example, 86.7\% of agencies reported using COTS AI to summarize long reports in 2025, compared with 39.0\% in 2024. Some of this rise is expected and typical considering the increasing adoption of AI in workplaces~\cite{baslandze2026artificial}. However, some of it likely follows from the new format, since an agency could answer yes as a blanket once rather than enter each use. More importantly, the move left the most widespread federal AI without a use-level account. For example, though 37 of 45 agencies named Microsoft Copilot in the 2025 file, the COTS report asked nothing about these uses' data, oversight, etc.

\subsection{Gating: Disclosure Tied to the Use Case Category}
Not all use cases had to answer all questions. In 2025, questions on training, testing, monitoring, and appeals applied only to deployed HI uses, and a written justification applied only to PHIN uses. That means an entry's classification determined which governance questions it had to answer, so a change in classification could add or remove those questions.

% Classification is thus where \emph{laws and policies} shaped the inventory most directly. We examine how agencies explained classifications, how HI entries answered the questions gated to them, and what motivations entries gave for AI at the classification boundary.

\subsubsection{Reclassifications Rarely Explained}
Most matched use cases maintained an equivalent classification between 2024 and 2025, but those that moved out of the highest category were rarely explained. Of the 272 matched uses that were rights- or safety-impacting in 2024, 228 (83.8\%) became HI in 2025, 16 became PHIN, and 28 became not HI. Of the 1,045 matched uses that were neither rights- nor safety-impacting in 2024, 902 (86.3\%) became not HI (Figure~\ref{fig:classification_transition} in Appendix). A move to PHIN required a justification, but a move directly to not-HI did not, and 18 of the 28 such uses, including all of the Department of Veterans Affairs (VA)'s, gave none. Thus, the reclassifications, which relieved these use cases from answering the governance questions, were the least explained.

Even where a justification was required, it rarely addressed the particular use. We identified four ways the PHIN use cases justified their classification. First, some simply \emph{recited the definition}. All 32 DOJ entries did this and used the same sentence: \emph{``Does not produce an output that serves as a principal basis for decisions or actions with legal, material, binding, or significant effect on any of the individuals or entities identified in OMB-25-21."} For example, LexisNexis Accurint (DOJ-0019) described its outputs identically in both years, yet moved from rights-impacting to PHIN with this sentence, and the impact assessment question it had left blank in 2024 no longer applied in 2025. Second, some justifications \emph{foregrounded the people who use the output}, for example, a Department of Homeland Security (DHS) use case (DHS-2729) stated that its results ``can only be utilized for investigative leads and cannot be utilized as a determining factor for any enforcement action", without saying how analysts weigh a match. Third, some \emph{narrowed the use}, as Customs and Border Protection's facial matching for airline check-in (DHS-2417) was ``an option for travelers, not a requirement." Finally, some entries \emph{deferred the determination}. For example, several Department of Energy entries stated ``In development stage and impact has not been determined." Overall, the 109 justifications contained only 69 distinct texts, suggesting that many reporting agencies might have assessed broad genres of use cases, but often not individual ones.

\subsubsection{Uneven Classifications and Risk Accounts}
Sometimes similar use cases received different classifications. For example, PATTERN (DOJ-0160), rights-impacting in both of its 2024 entries, appeared twice in 2025. While both entries output recidivism risk scores, the Federal Bureau of Prisons did not classify it as HI, whereas the Office of Justice Programs did. Similarly, Vivid iq (VA-24-3213) and Vivid E80/E90/E95 (VA-24-3254) were both rights- and safety-impacting in 2024 and reported the same output in 2025. However, only Vivid iq was HI in the latter year, and neither entry explained the reason behind such difference in their classifications.

Only HI use cases had to report potential impacts, safeguards, and human oversight, which serve as institutional means of managing the \emph{degree of uncertainty} in algorithmic decisions~\cite{metcalf2021algorithmic, green2022flaws}. However, 102 (44.9\%) of the deployed HI entries in 2025 left the question about potential impacts blank (94 of them from VA). The entries that answered described uncertainty in four types of accounts: deferring, proceduralizing, routing, and naming. \emph{Deferring} accounts said that impacts would be identified later, even for systems already in use: the AI-Assisted Resume Screening Tool (DHS-2556) stated that ``potential impacts will be identified during AI Impact Assessment." \emph{Proceduralizing} accounts simply named a governance process instead of explaining it: many DOJ entries answered every safeguard by saying this: \emph{``the case owner relied on DOJ AI governance practices to evaluate impacts and risks"}. \emph{Routing} accounts named an error and assigned it to people, sending it to a ``manual resolution queue" processed by ``authorized and trained personnel." Through such routing, human discretion is expected to absorb algorithmic uncertainty. Finally, \emph{naming} accounts identified errors without saying who would catch them: the Passenger Security Assessment Model (DHS-2389) reported risks include ``false positives and negatives [...] algorithmic bias may disproportionately target certain types of travelers and crossing behaviors'', but did not say who reviews a flagged traveler or how a missed detection would be noticed.

The 2025 inventory's only question about preparing the people who act on AI outputs was also applicable to HI use cases and was answered just as thinly. It asked whether the agency had established sufficient and periodic training for the AIs' operators, a matter of \emph{administration and training}. Of these, only 18.1\% reported established training, mostly from DHS and the Social Security Administration (SSA). Conversely, every DOJ entry reported training in progress, and every VA entry left the question blank. Even where training was established, entries did not say what operators were taught about interpreting or overriding AI outputs.

\subsubsection{Motivations at the Classification Boundary}
The classification asked whether a use affected rights or safety, but agencies justified AI mostly by efficiency, in terms of gains in time, cost, workload, accuracy, or consistency, whereas only a few use cases mentioned equity or fairness (Table~\ref{tab:benefit-keywords} in Appendix~\ref{app:benefits-keywords}). When entries mentioned the public, they usually presented efficiency as the route to it: faster processing would mean quicker access to services, benefits, or security. These \emph{value judgments} fell outside what the classification asked, so the question that gated disclosure and the reasons agencies gave for AI rarely met.

Even entries that acknowledged a tension between efficiency and rights/safety presented the two as compatible without saying how. Hurricane Score (DHS-2408) ``supports more consistent and efficient case reviews [...] while maintaining individualized assessments". However, it also conceded that an inaccurate score ``might lead to stricter or more lenient [...] requirements for an individual." Similarly, the Transportation Security Administration (TSA)'s PreCheck Touchless Identity Solution (DHS-345) promised ``increasing the speed of security checks while maintaining a high degree of safety for all passengers." Only a few use cases, such as TSA's Automated Target Recognition (DHS-131), named a rights-related goal such as ``reduc[ing] privacy concerns".
 
Entries also differed in the \emph{type of decision support} their outputs provided. Many scored, ranked, matched, or flagged cases. Such outputs act as a \emph{heuristic} for deciding which cases staff examine first and how closely. SSA's Quick Disability Determinations Model\footnote{Neither SSA model has an ID in 2024 or 2025.} identifies cases where a favorable determination ``is highly likely and medical evidence is readily available," so that these straightforward cases can be decided quickly. Conversely, the Pre-Effectuation Review and Targeted Denial Review models\footnotemark[\value{footnote}] flag claims with a ``high likelihood of error" for closer review. A second type of decision support helps enforce \emph{laws and policies} established by authorities. For example, the Biometrics Enrollment Tool (DHS-14) checks a fingerprint quality score against thresholds set ``to align with FBI\footnote{Federal Bureau of Investigation} specifications." Almost half of the triaging tools in the inventory were classified as HI. For the rest, a human still makes each final decision. However, by deciding which cases receive a quick decision and which receive scrutiny, it shapes outcomes before any person reviews the case.

% ===============================================================================================================
\subsection{Overwriting: Same Systems, Rewritten Answers}
Even when a use stayed in the inventory, and a question stayed the same, its answer could change in the following release, which replaced rather than amended the earlier account.

\subsubsection{Authorization-related Answers Reversed}
\begin{figure*}[t]
  \centering
  \scriptsize
  \begin{minipage}[t]{0.60\linewidth}\vspace{0pt}
    \centering
    \captionof{table}{Matched use cases' required response transition.}
    \label{tab:disclosure}
    \setlength{\tabcolsep}{2.5pt}
    \begin{tabular*}{\linewidth}{@{\extracolsep{\fill}}lrrrrrr@{}}
      \toprule
      Field & $n$ & Gain & Loss & \shortstack{Both\\answered} & \shortstack{Both\\blank} & Other \\
      \midrule
      Procurement/development$^{*}$ & 548 & 116 & 5 & 425 & 2 & 0 \\
      PII$^{\delta}$ use & 485 & 102 & 90 & 289 & 3 & 1 \\
      Data description$^{*}$ & 548 & 140 & 46 & 294 & 8 & 60 \\
      Custom-code status & 485 & 110 & 3 & 372 & 0 & 0 \\
      Authorization to Operate & 548 & 110 & 89 & 349 & 0 & 0 \\
      ATO system name & 231 & 0 & 19 & 212 & 0 & 0 \\
      \midrule
      Impact assessment$^{*}$ & 26 & 0 & 9 & 17 & 0 & 0 \\
      Independent review$^{*}$ & 26 & 0 & 9 & 17 & 0 & 0 \\
      Ongoing monitoring$^{*}$ & 26 & 1 & 9 & 16 & 0 & 0 \\
      Appeal/contestation$^{*}$ & 26 & 1 & 9 & 7 & 0 & 9 \\
      \bottomrule
    \end{tabular*}
    \par\smallskip
    {\raggedright\scriptsize $^{*}$Question scope, response options, or eligibility changed, so the comparison does not assume identical constructs. $^{\delta}$PII = Personally identifiable information.\par}
  \end{minipage}\hfill
  \begin{minipage}[t]{0.37\linewidth}\vspace{0pt}
    \centering
    \includegraphics[width=\linewidth]{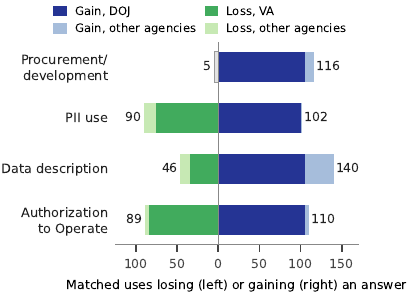}
    \caption{Gains and losses by agency for the four highlighted fields in Table~\ref{tab:disclosure}.}
    \label{fig:gains_losses_agency}
  \end{minipage}
\end{figure*}

We found that agencies often changed their answers about whether a use required an authority to operate (ATO), the security approval a federal information system needs. Both 2024 and 2025 inventories asked this question, and fewer than half of the entries required to answer it did so in either year (46.2\% in 2024 and 46.0\% in 2025). Among matched uses, these answers shifted at the agency level (Table~\ref{tab:disclosure}, Figure~\ref{fig:gains_losses_agency}): DOJ moved from leaving every answer blank to answering every entry, accounting for 105 of the 110 new answers, while VA did the opposite, accounting for 84 of the 89 new blanks. The Department of the Interior moved from mostly yes (75.4\%) to mostly no (88.4\%), with 52 matched uses changing from yes to no. Because an ATO could belong to a larger system, these answers could not show whether the AI itself had been reviewed, or whether a reversal reflected a new process or a different interpretation of the question by the responsible agency.

\subsubsection{More Responses but Less Specified Data}
Across releases, more entries answered the questions about the \emph{relevant data} behind algorithmic decision-making, but their answers offered little details. We focused on three questions: (i) data description, (ii) demographic variables used by an AI, and (iii) whether it used PII. Among entries required to answer, PII use rose from 14.5\% in 2024 to 22.8\% in 2025, and blank answers about demographic variables fell from 57.6\% to 34.5\%. However, fewer blanks did not mean more detail: only 65 entries in 2024 and 36 in 2025 named the demographic variables they used, most often age and sex/gender (Figure~\ref{fig:demographics} in App.~\ref{app:results}). Many answers offered a generic assurance.\footnote{E.g., \emph{``utilizes one or more of the demographic variables listed in compliance with all federal laws and agency regulations"}.}

For matched uses, whether an answer appeared or disappeared depended mostly on the reporting agency.\footnote{A blank answer in 2024 that became an explicit or substantive response in 2025 was a gain, and the reverse was a loss.} Of 548 matched uses, 140 (25.5\%) gained a data description, and 46 lost one; of 485 entries required to report PII, 102 gained an answer and 90 lost one (Table~\ref{tab:disclosure}). DOJ accounted for 105 of the data-description gains and 100 of the PII gains: it left the PII question blank for all of the 2024 entries that had to answer it, and blank for none in 2025. VA accounted for 34 of the data-description losses and 76 of the PII losses: it answered the PII question for nearly all of its 2024 entries and left it blank for all 159 in 2025. A detailed account could thus be overwritten by a blank. In 2024, VA's Care Assessment Needs Score, one of VA's AI use cases in 2024, described its data as ``Patient demographics, medical conditions, laboratory results, medications, health care utilization, social determinants of health, social history, vital signs" and named race or ethnicity, sex or gender, age, and socioeconomic status as model features, and explicitly stated that it did not involve PII. Its 2025 entry (VA-24-5501) left all three questions blank. A reader of the 2025 inventory would therefore know less about this system's data than a reader of the 2024 one, although nothing indicates that the system changed.

\subsubsection{Repositioning and Reframing AI's Roles}
Overwriting also changed where entries placed \emph{human discretion}, especially \emph{professional expertise}, relative to the AI. Because the inventories rarely asked who acts on AI outputs or with what expertise, its role had to be read from narratives, and could change between releases even when the questions did not. For example, in 2024, Nediser (VA-24-5464) compared its role to that of ``a Radiology resident who provides draft reports that the Attending Radiologist reviews," casting the system as a junior member of the profession. Its 2025 account replaced this analogy with a system that ``always requires human in the loop," moving expertise from the AI back to an unnamed human without any stated change to the system.

Such repositioning also rewrote accounts of an AI's uncertainty and recourse. Moreover, existing laws and authorities could override some OMB reporting requirements and withhold certain fields of some Department of Labor (DOL) entries from public reporting (e.g., DOL-26, DOL-44). In 2024, Hurricane Score (DHS-2408) identified two sources of uncertainty, false positives and bias, and addressed both ``through error analysis during model testing and model performance monitoring." Its 2025 account mentioned only false positives, repositioned the score ``as one of many factors" in case management, dropped country of origin from its demographic features, and changed its answer on contesting an outcome from an established fallback and escalation process to an appeal precluded by law. None of these revisions was explained, and a reader of the 2025 inventory alone would have no sign that a different account had existed a year earlier.%\clearpage
\section{Discussion}
\begin{figure*}[t]
    \centering
    \includegraphics[width=\linewidth]{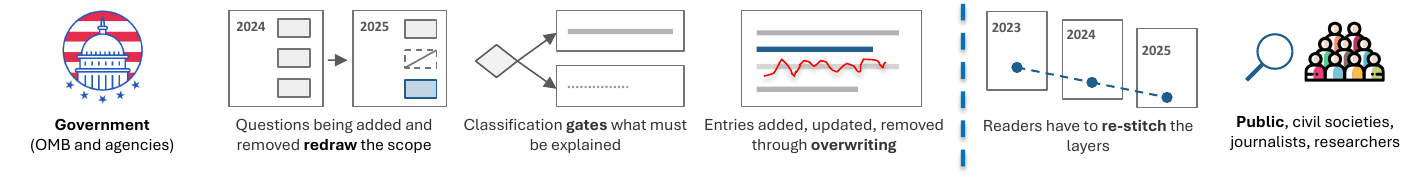}
    \caption{Mechanisms through which the US Federal AI Use Case Inventory is constructed (left) and construed (right). The government redraws, gates, and overwrites the inventory, and the public has to re-stitch it.}
    \label{fig:discussion}
\end{figure*}

Our study suggests that successive US Federal AI inventories offered visibility into the range of federal AI uses while poorly describing how any particular use was reclassified or described. We conceptualize this process through the metaphor of a palimpsest--a manuscript page scraped and written over, on which traces of the earlier writing remain partly legible~\cite{dillon2007palimpsest}. Following Dillon's use of ``palimpsestic" for this layering~\cite{dillon2005reinscribing}, we propose the term \emph{``palimpsestic accountability."} It means that the accountability an inventory supports depends not only on what each release discloses, but also on whether the layers of successive accounts can still be identified, separated, and connected. We identify three mechanisms, \emph{redrawing}, \emph{gating}, and \emph{overwriting}, through which the government prepared the releases, leaving the public to reconnect the layers through \emph{re-stitching} (see Figure~\ref{fig:discussion}). In the following subsections, we first theorize how these mechanisms wrote over the inventory and then discuss what they imply for public AI inventories.

\subsection{Theorizing Palimpsestic Accountability}
First, \emph{redrawing} changed the proverbial page itself. The 2024 inventory followed OMB memo M-24-10, whereas the 2025 inventory followed M-25-21, issued under the new administration~\cite{omb2024memo, omb2025memo}. The latter release removed the questions on resources even as entries justified AI by saved staff time, replaced free-text descriptions of techniques with fixed choices, and moved commercial AI products to a separate inventory detached from individual uses. Without a record of these changes, readers cannot tell whether an agency's practice changed or only the questions it was asked.
% The page also changed between releases: OMB guidance encourages agencies to update their inventories on an ongoing basis~\cite{omb2026guidance}, and DHS's July 2026 update retired four uses, including Hurricane Score~\cite{wilkinson2026dhs}, one of the systems we followed.
 
Second, \emph{gating} tied bureaucratic rules to accounts of algorithmic decision-making and associated human factors. In 2025, questions on training, testing, monitoring, and appeals applied only to deployed HI uses, and a justification only to PHIN use cases~\cite{omb2026guidance, ombegov2026federal}. A reclassification could thus remove a whole layer of governance disclosure without any answer being rewritten. Though classifications shape what institutions make visible~\cite{bowker2000sorting}, the inventory asked the least explanation where the classification moved the most: a use that moved directly from ``rights-impacting" to ``not-HI" owed no justification at all, and many required justifications repeated template text rather than addressing the particular use. Triage scores could also shift discretion upstream while the decision remained formally human and outside that gated disclosure. As policy researchers underscore: ``the determination that requires the most explanation is the one that comes with the least''~\cite{wilkinson2026dhs}.
 
Third, \emph{overwriting} rewrote entries one agency at a time, adding information in some and erasing it in others. Detailed answers in one release could become blanks in the next. Despite the concern of oversight policies often presuming capacities that reviewers lack~\cite{green2022flaws}, accounts of human discretion were overwritten in ways that obscured some systems' inaccuracies and biases, without any stated changes to the systems. Overwriting also continued between releases. As later news confirmed, DHS's July 2026 update retired four uses, including Hurricane Score~\cite{wilkinson2026dhs}, and after the April 2026 deadline, VA completed its risk-management responses for most uses and moved the others to categories that no longer required them~\cite{wilkinson2026risk}. However, the consolidated record curated and published by the OMB could not distinguish an unanswered question from an agency's pending update, or a completed answer from a reclassification that ended the requirement.
 
Finally, the inventory left \emph{re-stitching} to the public. With no identifiers carried from one release to the next, we pieced continuity together from titles, bureaus, and descriptions and manually reviewed fuzzy-matched candidate pairs. While traceability is often described as a principle for operationalizing accountability~\cite{kroll2021outlining}, under the current arrangement of how the OMB publishes the US Federal government's AI use case inventory, it became a task for the readers.
 
\subsection{Implications for Public AI Inventories}
A governance answer could disappear because the question was removed, because the use was reclassified out of it, because the agency left it blank, or because the use itself changed. The latest release would look the same in all four cases, though only the last reflects a change in what the government does with AI. To address this, we propose guidelines for palimpsestic accountability, with four recommendations for OMB and reporting Federal agencies. First, the inventory should preserve its own history. Each release should carry persistent public identifiers, explicit links for renamed, merged, split, or moved entries, and a change log of added, removed, or reworded questions, answer options, and eligibility rules, while revised answers and retired entries remain visible with their periods of being functional and reconciled with individual agencies' ongoing updates. Second, explanation requirements should follow the classification boundary: whenever a use crosses into or out of a rights-, safety-, or HI category, its entry should state the previous and new categories, the responsible office, the evidence considered, and whether the system, the criteria, or the record changed. Third, public files should distinguish a blank required answer from a pending update, an ended requirement, an answer withheld by law, and an explicit \emph{not applicable}, and should mark statements shared across an agency's entries as such. Fourth, federal agencies should document human oversight as a set of conditions, including when review occurs, what evidence reviewers receive, whether they can reject an output, what training and time they have, and how affected people can contest an outcome. Overall, palimpsestic accountability aims to make each revision legible as a recorded change to an earlier account and not as an isolated account.

These recommendations also apply beyond the US. The EU's DSA Transparency Database has been inconsistent across reporting periods~\cite{groesch2025big, trujillo2026disarranged}, and German register platforms neither keep earlier versions nor require updates~\cite{peljto2026are}. Canada's second AI Register, due by the end of 2026, will replace its 2025 minimum viable product (MVP)~\cite{gc2026nap, gc2025register}. Without persistent identifiers or a record of changed fields, we expect the same mechanisms: scopes redrawn beyond the MVP, disclosure gated by updated classifications, and MVP entries overwritten. As registers accumulate releases, palimpsestic accountability calls for reading AI use cases as traceable series, and not snapshots.%\clearpage
\section{Conclusion}
This paper traced the US Federal AI Use Case Inventory across three releases (2023--2025), focusing on how the record is written over successive reporting cycles. We showed that reclassifications and the conditions of human involvement were rarely explained in case-specific terms. To address how each release writes over the last, leaving the public to reconnect its accounts, we proposed \emph{palimpsestic accountability} and offered recommendations. While our analysis relies only on documentary accounts, future research engaging civil servants involved in these AI uses and their affected communities could explain how AI use cases are reclassified or safeguards change.
% \clearpage
% \input{sections/endmatters}
% \clearpage

%TC:ignore
\bibliographystyle{ACM-Reference-Format}
\bibliography{sample-base}

@misc{ombegov2026federal,
  title={2025 Federal Agency AI Use Case Inventory},
  author={Office of the Federal Chief Information Officer},
  howpublished={\url{https://github.com/ombegov/2025-Federal-Agency-AI-Use-Case-Inventory}},
  year={2026},
  note={[Accessed: April 25, 2026]}
}

@misc{omb2026guidance,
  title={Guidance on 2025 Agency Artificial Intelligence Reporting},
  author={Office of Management and Budget},
  howpublished={\url{https://fedscoop.com/wp-content/uploads/sites/5/2026/04/Guidance-on-2025-Agency-Artificial-Intelligence-Reporting-FINAL-3.pdf}},
  year={2026},
  note={[Accessed: April 25, 2026]}
}

@misc{omb2025memo,
  title={MEMORANDUM FOR THE HEADS OF EXECUTIVE DEPARTMENTS AND AGENCIES},
  author={Office of Management and Budget},
  howpublished={\url{https://www.whitehouse.gov/wp-content/uploads/2025/02/M-25-21-Accelerating-Federal-Use-of-AI-through-Innovation-Governance-and-Public-Trust.pdf}},
  year={2025},
  note={[Accessed: June 12, 2026]}
}

@book{bowker2000sorting,
  title={Sorting things out: Classification and its consequences},
  author={Bowker, Geoffrey C and Star, Susan Leigh},
  year={2000},
  publisher={MIT press}
}

@article{mcdonald2019reliability,
  title={Reliability and inter-rater reliability in qualitative research: Norms and guidelines for CSCW and HCI practice},
  author={McDonald, Nora and Schoenebeck, Sarita and Forte, Andrea},
  journal={Proceedings of the ACM on human-computer interaction},
  volume={3},
  number={CSCW},
  pages={1--23},
  year={2019},
  publisher={ACM New York, NY, USA}
}

@article{saxena2021framework,
  title={A framework of high-stakes algorithmic decision-making for the public sector developed through a case study of child-welfare},
  author={Saxena, Devansh and Badillo-Urquiola, Karla and Wisniewski, Pamela J and Guha, Shion},
  journal={Proceedings of the ACM on Human-Computer Interaction},
  volume={5},
  number={CSCW2},
  pages={1--41},
  year={2021},
  publisher={ACM New York, NY, USA}
}

@inproceedings{das2026bureaucratic,
  author    = {Das, Dipto and Tessono, Christelle and Ahmed, Syed Ishtiaque and Guha, Shion},
  title     = {Bureaucratic Silences: What the Canadian AI Register Reveals, Omits, and Obscures},
  year      = {2026},
  month     = jun,
  isbn      = {979-8-4007-2596-8/2026/06},
  publisher = {Association for Computing Machinery},
  address   = {New York, NY, USA},
  url       = {https://doi.org/10.1145/3805689.3812336},
  doi       = {10.1145/3805689.3812336},
  booktitle = {Proceedings of the 2026 ACM Conference on Fairness, Accountability, and Transparency (FAccT '26)},
  location  = {Montr\'eal, Canada},
  series    = {FAccT '26}
}

@article{ananny2018seeing,
  title={Seeing without knowing: Limitations of the transparency ideal and its application to algorithmic accountability},
  author={Ananny, Mike and Crawford, Kate},
  journal={new media \& society},
  volume={20},
  number={3},
  pages={973--989},
  year={2018},
  publisher={SAGE Publications Sage UK: London, England}
}

@inproceedings{kizilcec2016much,
  title={How much information? Effects of transparency on trust in an algorithmic interface},
  author={Kizilcec, Ren{\'e} F},
  booktitle={Proceedings of the 2016 CHI conference on human factors in computing systems},
  pages={2390--2395},
  year={2016}
}

@inproceedings{das2026asymmetries,
  author    = {Das, Dipto and Tamura, Matthew and Ahmed, Syed Ishtiaque and Guha, Shion},
  title     = {``Is This Not Enough?'': Asymmetries in Institutional Accountability and Collective Sensemaking in the Case of Canada's Algorithmic Visa Triage System},
  booktitle = {Proceedings of the ACM SIGCAS/SIGCHI Conference on Computing and Sustainable Societies},
  series    = {COMPASS '26},
  year      = {2026},
  date      = {2026-07-27/2026-07-31},
  location  = {Virtual Event},
  publisher = {Association for Computing Machinery},
  address   = {New York, NY, USA},
  url       = {https://doi.org/10.1145/3811242.3819106},
  doi       = {10.1145/3811242.3819106}
}

@inproceedings{mitchell2019model,
  title={Model cards for model reporting},
  author={Mitchell, Margaret and Wu, Simone and Zaldivar, Andrew and Barnes, Parker and Vasserman, Lucy and Hutchinson, Ben and Spitzer, Elena and Raji, Inioluwa Deborah and Gebru, Timnit},
  booktitle={Proceedings of the conference on fairness, accountability, and transparency},
  pages={220--229},
  year={2019}
}

@article{gebru2021datasheets,
  title={Datasheets for datasets},
  author={Gebru, Timnit and Morgenstern, Jamie and Vecchione, Briana and Vaughan, Jennifer Wortman and Wallach, Hanna and III, Hal Daum{\'e} and Crawford, Kate},
  journal={Communications of the ACM},
  volume={64},
  number={12},
  pages={86--92},
  year={2021},
  publisher={ACM New York, NY, USA}
}

@inproceedings{kroll2021outlining,
  title={Outlining traceability: A principle for operationalizing accountability in computing systems},
  author={Kroll, Joshua A},
  booktitle={Proceedings of the 2021 ACM Conference on fairness, accountability, and transparency},
  pages={758--771},
  year={2021}
}

@inproceedings{eyert2023rethinking,
  title={Rethinking transparency as a communicative constellation},
  author={Eyert, Florian and Lopez, Paola},
  booktitle={Proceedings of the 2023 ACM Conference on Fairness, Accountability, and Transparency},
  pages={444--454},
  year={2023}
}

@inproceedings{norval2022disclosure,
  title={Disclosure by Design: Designing information disclosures to support meaningful transparency and accountability},
  author={Norval, Chris and Cornelius, Kristin and Cobbe, Jennifer and Singh, Jatinder},
  booktitle={Proceedings of the 2022 ACM Conference on Fairness, Accountability, and Transparency},
  pages={679--690},
  year={2022}
}

@inproceedings{poirier2022accountable,
  title={Accountable data: The politics and pragmatics of disclosure datasets},
  author={Poirier, Lindsay},
  booktitle={Proceedings of the 2022 ACM Conference on Fairness, Accountability, and Transparency},
  pages={1446--1456},
  year={2022}
}

@misc{gc2025register,
  author={Government of Canada},
  title={Government of Canada AI Register (Minimum Viable Product) - GC AI Register (MVP)},
  year={2025},
  howpublished={\url{https://open.canada.ca/data/en/dataset/fcbc0200-79ba-4fa4-94a6-00e32facea6b/resource/369f6f34-148a-42ed-b581-8c164e941a89}},
  note={[Accessed: December 16, 2025]}
}

@article{poirier2021reading,
  title={Reading datasets: Strategies for interpreting the politics of data signification},
  author={Poirier, Lindsay},
  journal={Big Data \& Society},
  volume={8},
  number={2},
  pages={20539517211029322},
  year={2021},
  publisher={SAGE Publications Sage UK: London, England}
}

@article{denton2021genealogy,
  title={On the genealogy of machine learning datasets: A critical history of ImageNet},
  author={Denton, Emily and Hanna, Alex and Amironesei, Razvan and Smart, Andrew and Nicole, Hilary},
  journal={Big Data \& Society},
  volume={8},
  number={2},
  pages={20539517211035955},
  year={2021},
  publisher={SAGE Publications Sage UK: London, England}
}

@inproceedings{kaushal2024automated,
  title={Automated transparency: A legal and empirical analysis of the digital services act transparency database},
  author={Kaushal, Rishabh and Van De Kerkhof, Jacob and Goanta, Catalina and Spanakis, Gerasimos and Iamnitchi, Adriana},
  booktitle={Proceedings of the 2024 ACM Conference on Fairness, Accountability, and Transparency},
  pages={1121--1132},
  year={2024}
}

@article{nieuwenhuizen2024algorithm,
  title={Algorithm Registers: A Box-Ticking Exercise or Meaningful Tool for Transparency?},
  author={Nieuwenhuizen, Esther},
  journal={Information Polity},
  volume={29},
  number={4},
  pages={415--433},
  year={2024},
  publisher={SAGE Publications Sage UK: London, England}
}

@article{correa2023worldwide,
  title={Worldwide AI ethics: A review of 200 guidelines and recommendations for AI governance},
  author={Corr{\^e}a, Nicholas Kluge and Galv{\~a}o, Camila and Santos, James William and Del Pino, Carolina and Pinto, Edson Pontes and Barbosa, Camila and Massmann, Diogo and Mambrini, Rodrigo and Galv{\~a}o, Luiza and Terem, Edmund and others},
  journal={Patterns},
  volume={4},
  number={10},
  year={2023},
  publisher={Elsevier}
}

@article{haataja2020public,
  title={Public AI Registers: Realising AI transparency and civic participation in government use of AI},
  author={Haataja, M and van de Fliert, L and Rautio, P},
  journal={Whitepaper Version},
  volume={1},
  year={2020}
}

@article{groesch2025big,
  title={Big data, small answers: How the DSA Transparency Database falls short of its regulatory objectives},
  author={Groesch, Samuel and Birrer, Alena and Just, Natascha and Saurwein, Florian},
  journal={Telecommunications Policy},
  pages={103088},
  year={2025},
  publisher={Elsevier}
}

@inproceedings{raji2020closing,
  title={Closing the AI accountability gap: Defining an end-to-end framework for internal algorithmic auditing},
  author={Raji, Inioluwa Deborah and Smart, Andrew and White, Rebecca N and Mitchell, Margaret and Gebru, Timnit and Hutchinson, Ben and Smith-Loud, Jamila and Theron, Daniel and Barnes, Parker},
  booktitle={Proceedings of the 2020 conference on fairness, accountability, and transparency},
  pages={33--44},
  year={2020}
}

@inproceedings{watkins2021governing,
  title={Governing algorithmic systems with impact assessments: Six observations},
  author={Watkins, Elizabeth Anne and Moss, Emanuel and Metcalf, Jacob and Singh, Ranjit and Elish, Madeleine Clare},
  booktitle={Proceedings of the 2021 AAAI/ACM Conference on AI, Ethics, and Society},
  pages={1010--1022},
  year={2021}
}

@inproceedings{moon2025datafication,
  title={The Datafication of Care in Public Homelessness Services},
  author={Moon, Erina Seh-Young and Saxena, Devansh and Das, Dipto and Guha, Shion},
  booktitle={Proceedings of the 2025 CHI Conference on Human Factors in Computing Systems},
  pages={1--16},
  year={2025}
}

@inproceedings{saxena2023rethinking,
  title={Rethinking" Risk" in Algorithmic Systems Through A Computational Narrative Analysis of Casenotes in Child-Welfare},
  author={Saxena, Devansh and Moon, Erina Seh-Young and Chaurasia, Aryan and Guan, Yixin and Guha, Shion},
  booktitle={Proceedings of the 2023 CHI conference on human factors in computing systems},
  pages={1--19},
  year={2023}
}

@misc{gpai2024algorithmic,
  title={Algorithmic Transparency in the Public Sector: A state-of-the-art report ofalgorithmic transparency instruments},
  author={Global Partnership on Artificial Intelligence},
  howpublished={\url{https://wp.oecd.ai/app/uploads/2025/05/algorithmic-transparency-in-the-public-sector.pdf}},
  year={2024},
  note={[Accessed: Dec 28, 2025]}
}

@book{eubanks2018automating,
  title={Automating inequality: How high-tech tools profile, police, and punish the poor},
  author={Eubanks, Virginia},
  year={2018},
  publisher={Macmillan+ ORM}
}

@article{scheuerman2025data,
  title={How Data Workers Shape Datasets: The Role of Positionality in Data Collection and Annotation for Computer Vision},
  author={Scheuerman, Morgan Klaus and Woodruff, Allison and Brubaker, Jed R},
  journal={Proceedings of the ACM on Human-Computer Interaction},
  volume={9},
  number={7},
  pages={1--42},
  year={2025},
  publisher={ACM New York, NY, USA}
}

@inproceedings{mcconvey2024not,
  title={" This is not a data problem": Algorithms and Power in Public Higher Education in Canada},
  author={McConvey, Kelly and Guha, Shion},
  booktitle={Proceedings of the 2024 CHI Conference on Human Factors in Computing Systems},
  pages={1--14},
  year={2024}
}

@inproceedings{winecoff2025improving,
  title={Improving governance outcomes through AI documentation: Bridging theory and practice},
  author={Winecoff, Amy and Bogen, Miranda},
  booktitle={Proceedings of the 2025 CHI Conference on Human Factors in Computing Systems},
  pages={1--18},
  year={2025}
}

@article{redden2022automating,
  title={Automating public services: Learning from cancelled systems},
  author={Redden, Joanna and Brand, Jessica and Sander, Ina and Warne, Harry and Grant, A and White, D},
  journal={Cardiff, UK: Data Justice Lab, Cardiff University},
  volume={30},
  year={2022}
}

@misc{sieber2026building,
  title={Building AI Governance in Municipalities from the Ground Up},
  author={Sieber, Ren{\'e}e and Brandusescu, Ana and van Geuns, Jonathan},
  year={2026},
  publisher={University of Toronto, School of Cities: Toronto, ON, Canada}
}

@article{sengupta2022algorithm,
  title={Algorithm registers as an infrastructural element for transparent government},
  author={Sengupta, Ushnish},
  journal={Artificial intelligence in the city: Building civic engagement and public trust},
  pages={68},
  year={2022}
}

@article{pi2026understanding,
  title={Understanding the Role of Algorithm Registers in AI Governance Through Comparative Analysis of China and the UK},
  author={Pi, Yulu and Li, Wenlong and Singh, Jatinder},
  journal={arXiv preprint arXiv:2606.00035},
  year={2026}
}

@inproceedings{lawrence2023bureaucratic,
  title={The bureaucratic challenge to AI governance: An empirical assessment of implementation at US federal agencies},
  author={Lawrence, Christie and Cui, Isaac and Ho, Daniel},
  booktitle={Proceedings of the 2023 AAAI/ACM Conference on AI, Ethics, and Society},
  pages={606--652},
  year={2023}
}

@article{jobin2019global,
  title={The global landscape of AI ethics guidelines},
  author={Jobin, Anna and Ienca, Marcello and Vayena, Effy},
  journal={Nature machine intelligence},
  volume={1},
  number={9},
  pages={389--399},
  year={2019},
  publisher={Nature Publishing Group UK London}
}

@article{green2022flaws,
  title={The flaws of policies requiring human oversight of government algorithms},
  author={Green, Ben},
  journal={Computer Law \& Security Review},
  volume={45},
  pages={105681},
  year={2022},
  publisher={Elsevier}
}

@misc{ogp2021building,
  title={Building Public Algorithm Registers: Lessons Learned from the French Approach},
  author={Open Government Partnership},
  howpublished={\url{https://www.opengovpartnership.org/stories/building-public-algorithm-registers-lessons-learned-from-the-french-approach/}},
  year={2021},
  note={[Accessed: June 11, 2026]}
}

@misc{president2020promoting,
  title={Promoting the Use of Trustworthy Artificial Intelligence in the Federal Government},
  author={Executive Office of the President},
  howpublished={\url{https://www.federalregister.gov/documents/2020/12/08/2020-27065/promoting-the-use-of-trustworthy-artificial-intelligence-in-the-federal-government}},
  year={2020},
  note={[Accessed: June 12, 2026]}
}

@misc{peters2021advancing,
  title={Advancing American AI Act},
  author={Peters, Gary C.},
  howpublished={\url{https://www.congress.gov/bill/117th-congress/senate-bill/1353/text}},
  year={2021},
  note={[Accessed: June 12, 2026]}
}

@inproceedings{lee2006human,
  title={The human infrastructure of cyberinfrastructure},
  author={Lee, Charlotte P and Dourish, Paul and Mark, Gloria},
  booktitle={Proceedings of the 2006 20th anniversary conference on Computer supported cooperative work},
  pages={483--492},
  year={2006}
}

@article{jaro1989advances,
  title={Advances in record-linkage methodology as applied to matching the 1985 census of Tampa, Florida},
  author={Jaro, Matthew A},
  journal={Journal of the American Statistical association},
  volume={84},
  number={406},
  pages={414--420},
  year={1989},
  publisher={Taylor \& Francis}
}

@article{crouse2016implementing,
  title={On implementing 2D rectangular assignment algorithms},
  author={Crouse, David F},
  journal={IEEE Transactions on Aerospace and Electronic Systems},
  volume={52},
  number={4},
  pages={1679--1696},
  year={2016},
  publisher={IEEE}
}

@book{dillon2007palimpsest,
  author = {Dillon, Sarah},
  title = {The Palimpsest: Literature, Criticism, Theory},
  publisher = {Continuum},
  address = {London and New York},
  year = {2007}
}

@article{dillon2005reinscribing,
  author = {Dillon, Sarah},
  title = {Reinscribing {D}e {Q}uincey's Palimpsest: The Significance of the Palimpsest in Contemporary Literary and Cultural Studies},
  journal = {Textual Practice},
  volume = {19},
  number = {3},
  pages = {243--263},
  year = {2005}
}

@misc{aldane2026number,
  author = {Aldane, Jack},
  title = {Number of {AI} Use Cases in the {US} Federal Government More Than Doubles in a Year, Official Figures Show},
  howpublished = {Global Government Forum},
  year = {2026},
  howpublished = {\url{https://www.globalgovernmentforum.com/number-of-ai-use-cases-in-the-us-federal-government-more-than-doubles-in-a-year-official-figures-show/}},
  note = {[Accessed: September 28, 2026]}
}

@misc{kelley2026agencies,
  author = {Kelley, Alexandra},
  title = {Agencies Report over 3,000 {AI} Use Cases in 2025},
  year = {2026},
  howpublished = {\url{https://www.nextgov.com/artificial-intelligence/2026/04/agencies-report-over-3000-ai-use-cases-2025/412898/}},
  note = {[Accessed: September 28, 2026]}
}

@misc{wilkinson2026risk,
  author = {Wilkinson, Lindsey},
  title = {{AI} Risk Management Deadline Hits Federal Agencies. {N}ot All Were Ready.},
  year = {2026},
  howpublished = {\url{https://fedscoop.com/federal-agencies-ai-inventory-risk-management-deadline/}},
  note = {[Accessed: September 28, 2026]}
}

@misc{wilkinson2026dhs,
  author = {Wilkinson, Lindsey},
  title = {Homeland Security's Updated {AI} Inventory Raises More Questions than It Answers},
  year = {2026},
  howpublished = {\url{https://fedscoop.com/dhs-updated-ai-inventory-risk-management-high-impact/}},
  note = {[Accessed: September 28, 2026]}
}

@misc{wirtschafter2025assessing,
  author = {Wirtschafter, Valerie},
  title = {{A}ssessing the state of {AI} adoption across the federal government},
  howpublished = {\url{https://www.brookings.edu/articles/assessing-the-state-of-ai-adoption-across-the-federal-government/}},
  year = {2025},
  note = {[Accessed: September 29, 2026]},
}

@inproceedings{kawakami2022improving,
  author    = {Kawakami, Anna and Sivaraman, Venkatesh and Cheng, Hao-Fei and Stapleton, Logan and Cheng, Yanghuidi and Qing, Diana and Perer, Adam and Wu, Zhiwei Steven and Zhu, Haiyi and Holstein, Kenneth},
  title     = {Improving Human-{AI} Partnerships in Child Welfare: Understanding Worker Practices, Challenges, and Desires for Algorithmic Decision Support},
  booktitle = {Proceedings of the 2022 CHI Conference on Human Factors in Computing Systems},
  series    = {CHI '22},
  pages     = {1--18},
  year      = {2022},
  publisher = {ACM},
  address   = {New Orleans, LA, USA},
  doi       = {10.1145/3491102.3517439}
}

@article{bowen2006grounded,
  title={Grounded theory and sensitizing concepts},
  author={Bowen, Glenn A},
  journal={International journal of qualitative methods},
  volume={5},
  number={3},
  pages={12--23},
  year={2006},
  publisher={SAGE Publications, Sage CA: Los Angeles, CA}
}

@article{nolte2020support,
  title={How to support newcomers in scientific hackathons-an action research study on expert mentoring},
  author={Nolte, Alexander and Hayden, Linda Bailey and Herbsleb, James D},
  journal={Proceedings of the ACM on Human-Computer Interaction},
  volume={4},
  number={CSCW1},
  pages={1--23},
  year={2020},
  publisher={ACM, New York, NY, USA}
}

@article{kiene2026relational,
  title={The relational origins of rules in online communities},
  author={Kiene, Charles and Hwang, Sohyeon and Colglazier, Carl and TeBlunthuis, Nathan and Shaw, Aaron and Hill, Benjamin Mako},
  journal={Proceedings of the ACM on Human-Computer Interaction},
  volume={10},
  number={6},
  pages={1--34},
  year={2026},
  publisher={ACM, New York, NY}
}

@article{goldkuhl2024aspects,
  title={Aspects of E-Government Interoperation: Case Study Research for Discovery Beyond Interoperability as Layered Capability},
  author={Goldkuhl, G{\"o}ran},
  journal={ACM SIGMIS Database: the DATABASE for Advances in Information Systems},
  volume={55},
  number={4},
  pages={53--79},
  year={2024},
  publisher={ACM, New York, NY, USA}
}

@article{selten2023just,
  author  = {Selten, Friso and Robeer, Marcel and Grimmelikhuijsen, Stephan},
  title   = {`Just like {I} thought': Street-level Bureaucrats Trust {AI} Recommendations if They Confirm Their Professional Judgment},
  journal = {Public Administration Review},
  volume  = {83},
  number  = {2},
  pages   = {263--278},
  year    = {2023},
  doi     = {10.1111/puar.13602}
}

@article{liang2024systematic,
  author  = {Liang, Weixin and Rajani, Nazneen and Yang, Xinyu and Ozoani, Ezinwanne and Wu, Eric and Chen, Yiqun and Smith, Daniel Scott and Zou, James},
  title   = {Systematic Analysis of 32,111 {AI} Model Cards Characterizes Documentation Practice in {AI}},
  journal = {Nature Machine Intelligence},
  volume  = {6},
  pages   = {744--753},
  year    = {2024},
  doi     = {10.1038/s42256-024-00857-z},
  note    = {Preprint: arXiv:2402.05160, ``What's documented in AI? Systematic Analysis of 32K AI Model Cards''}
}

@inproceedings{metcalf2021algorithmic,
  author    = {Metcalf, Jacob and Moss, Emanuel and Watkins, Elizabeth Anne and Singh, Ranjit and Elish, Madeleine Clare},
  title     = {Algorithmic Impact Assessments and Accountability: The Co-construction of Impacts},
  booktitle = {Proceedings of the 2021 ACM Conference on Fairness, Accountability, and Transparency},
  series    = {FAccT '21},
  pages     = {735--746},
  year      = {2021},
  publisher = {ACM},
  doi       = {10.1145/3442188.3445935}
}

@article{selbst2021institutional,
  author  = {Selbst, Andrew D.},
  title   = {An Institutional View of Algorithmic Impact Assessments},
  journal = {Harvard Journal of Law \& Technology},
  volume  = {35},
  number  = {1},
  pages   = {117},
  year    = {2021},
  url     = {https://papers.ssrn.com/sol3/papers.cfm?abstract_id=3867634}
}

@article{kaminski2023regulating,
  author  = {Kaminski, Margot E.},
  title   = {Regulating the Risks of {AI}},
  journal = {Boston University Law Review},
  volume  = {103},
  number  = {5},
  pages   = {1347--1410},
  year    = {2023},
  url     = {https://www.bu.edu/bulawreview/files/2023/11/KAMINSKI.pdf}
}

@inproceedings{golpayegani2023high,
  author    = {Golpayegani, Delaram and Pandit, Harshvardhan J. and Lewis, Dave},
  title     = {To Be High-Risk, or Not To Be---Semantic Specifications and Implications of the {AI} Act's High-Risk {AI} Applications and Harmonised Standards},
  booktitle = {Proceedings of the 2023 ACM Conference on Fairness, Accountability, and Transparency},
  series    = {FAccT '23},
  pages     = {905--915},
  year      = {2023},
  publisher = {ACM},
  doi       = {10.1145/3593013.3594050}
}

@article{friedler2025omb,
  author  = {Friedler, Sorelle A. and Selbst, Andrew D.},
  title   = {The {OMB} Artificial Intelligence Memoranda},
  journal = {Berkeley Technology Law Journal},
  volume  = {40},
  number  = {4},
  pages   = {1237--1268},
  year    = {2025}
}

@inproceedings{luccioni2022framework,
  author    = {Luccioni, Alexandra Sasha and Corry, Frances and Sridharan, Hamsini and Ananny, Mike and Schultz, Jason and Crawford, Kate},
  title     = {A Framework for Deprecating Datasets: Standardizing Documentation, Identification, and Communication},
  booktitle = {Proceedings of the 2022 ACM Conference on Fairness, Accountability, and Transparency},
  series    = {FAccT '22},
  pages     = {199--212},
  year      = {2022},
  publisher = {ACM},
  address   = {Seoul, Republic of Korea},
  doi       = {10.1145/3531146.3533086}
}

@misc{abraham2026silent,
  author        = {Abraham, Sophia and Bucknall, Ben},
  title         = {Silent Updates: Measuring and Closing the Post-Deployment Disclosure Gap},
  year          = {2026},
  eprint        = {2608.11803},
  archivePrefix = {arXiv},
  primaryClass  = {cs.CY},
  url           = {https://arxiv.org/abs/2608.11803}
}

@inproceedings{shahi2025year,
  author    = {Shahi, Gautam Kishore and Tessa, Benedetta and Trujillo, Amaury and Cresci, Stefano},
  title     = {A Year of the {DSA} Transparency Database: What It (Does Not) Reveal About Platform Moderation During the 2024 {European Parliament} Election},
  booktitle = {Workshop Proceedings of the 19th International AAAI Conference on Web and Social Media (ICWSM)},
  year      = {2025},
  note      = {arXiv:2504.06976},
  url       = {https://workshop-proceedings.icwsm.org/pdf/2025_08.pdf}
}

@misc{trujillo2026disarranged,
  author        = {Trujillo, Amaury and Tessa, Benedetta and Cresci, Stefano},
  title         = {Disarranged Harmonization of Transparency Reporting by Social Media Platforms Under the {Digital Services Act}},
  year          = {2026},
  eprint        = {2605.17655},
  archivePrefix = {arXiv},
  url           = {https://arxiv.org/abs/2605.17655}
}

@article{floridi2020artificial,
  author  = {Floridi, Luciano},
  title   = {Artificial Intelligence as a Public Service: Learning from {Amsterdam} and {Helsinki}},
  journal = {Philosophy \& Technology},
  volume  = {33},
  number  = {4},
  pages   = {541--546},
  year    = {2020},
  doi     = {10.1007/s13347-020-00434-3}
}

@inproceedings{lapostol2023algorithmic,
  author    = {Lapostol Piderit, Jos{\'e} Pablo and Garrido Iglesias, Romina and Hermosilla Cornejo, Mar{\'i}a Paz},
  title     = {Algorithmic Transparency from the South: Examining the State of Algorithmic Transparency in {Chile}'s Public Administration Algorithms},
  booktitle = {Proceedings of the 2023 ACM Conference on Fairness, Accountability, and Transparency},
  series    = {FAccT '23},
  year      = {2023},
  pages     = {227--235},
  publisher = {ACM},
  doi       = {10.1145/3593013.3593991}
}

@inproceedings{zilka2023exploring,
  author    = {Zilka, Miri and Ashurst, Carolyn and Chambers, Luke and Goodman, Ellen P. and Ugwudike, Pamela and Oswald, Marion},
  title     = {Exploring Police Perspectives on Algorithmic Transparency: A Qualitative Analysis of Police Interviews in the {UK}},
  booktitle = {Proceedings of the 3rd ACM Conference on Equity and Access in Algorithms, Mechanisms, and Optimization},
  series    = {EAAMO '23},
  year      = {2023},
  pages     = {1--19},
  publisher = {ACM},
  doi       = {10.1145/3617694.3623246}
}

@article{cath2022dutch,
  author  = {Cath, Corinne and Jansen, Fieke},
  title   = {Dutch Comfort: The Limits of {AI} Governance through Municipal Registers},
  journal = {Techn{\'e}: Research in Philosophy and Technology},
  volume  = {26},
  number  = {3},
  pages   = {395--412},
  year    = {2022},
  doi     = {10.5840/techne202323172}
}

@inproceedings{schuitemaker2024realizing,
  author    = {Schuitemaker, Nena and van Vliet, Martijn and Brinkkemper, Sjaak and van de Weerd, Inge and Espa{\~n}a, Sergio},
  title     = {Realizing the Accountability of Algorithms in the Public Sector: A Reference Method for Managing Algorithm Registers},
  booktitle = {Enterprise, Business-Process and Information Systems Modeling (BPMDS 2024, EMMSAD 2024)},
  series    = {Lecture Notes in Business Information Processing},
  volume    = {511},
  pages     = {325--340},
  year      = {2024},
  publisher = {Springer},
  doi       = {10.1007/978-3-031-61007-3_24}
}

@misc{peljto2026are,
  author        = {Peljto, Iman and Heilmann, Xenia and Cerrato, Mattia},
  title         = {Are Algorithm Registers Transparent? Perspectives from {Germany}},
  year          = {2026},
  eprint        = {2606.02347},
  archivePrefix = {arXiv},
  primaryClass  = {cs.CY},
  url           = {https://arxiv.org/abs/2606.02347}
}

@misc{detroya2026coconstructing,
  author        = {de Troya, {\'I}{\~n}igo and Enbergs, Maurus and Doorn, Neelke and Dobbe, Roel},
  title         = {Co-constructing Sociotechnical {AI} Governance: Participatory System Mapping Using Algorithm Registers},
  year          = {2026},
  eprint        = {2608.12166},
  archivePrefix = {arXiv},
  primaryClass  = {cs.CY},
  url           = {https://arxiv.org/abs/2608.12166}
}

@inproceedings{das2026global,
  author    = {Das, Dipto and Guha, Shion},
  title     = {A Global Comparison of Schemas, Transparency, and Interoperability in Public-Sector {AI} Registers and Inventories},
  booktitle = {Proceedings of the Inaugural ACM Conference on Digital Transformation},
  series    = {DXConf '26},
  year      = {2026},
  address   = {Ann Arbor, MI, USA},
  publisher = {Association for Computing Machinery},
  doi       = {10.1145/3847238.3849037}
}

@techreport{engstrom2020government,
  author      = {Engstrom, David Freeman and Ho, Daniel E. and Sharkey, Catherine M. and Cu{\'e}llar, Mariano-Florentino},
  title       = {Government by Algorithm: Artificial Intelligence in Federal Administrative Agencies},
  institution = {Administrative Conference of the United States},
  type        = {Report submitted to the Administrative Conference of the United States},
  year        = {2020},
  month       = feb,
  url         = {https://www.acus.gov/document/government-algorithm-artificial-intelligence-federal-administrative-agencies},
}

@misc{omb2024memo,
  author       = {{Office of Management and Budget}},
  title        = {Advancing Governance, Innovation, and Risk Management for Agency Use of Artificial Intelligence},
  howpublished = {Memorandum M-24-10},
  year         = {2024},
  month        = mar,
  note         = {March 28, 2024},
  url          = {https://www.whitehouse.gov/wp-content/uploads/2024/03/M-24-10-Advancing-Governance-Innovation-and-Risk-Management-for-Agency-Use-of-Artificial-Intelligence.pdf}
}

@techreport{gao2023artificial,
  author      = {{U.S. Government Accountability Office}},
  title       = {Artificial Intelligence: Agencies Have Begun Implementation but Need to Complete Key Requirements},
  institution = {U.S. Government Accountability Office},
  number      = {GAO-24-105980},
  year        = {2023},
  month       = dec,
  note        = {Published and publicly released December 12, 2023},
  url         = {https://www.gao.gov/products/gao-24-105980}
}

@techreport{wang2025assessing,
  author      = {Wang, Jennifer and Suzgun, Mirac and Meinhardt, Caroline and Zhang, Daniel and Nowacki, Kazia and Ho, Daniel E.},
  title       = {Assessing the Implementation of Federal {AI} Leadership and Compliance Mandates},
  institution = {Stanford HAI and RegLab},
  type        = {White Paper},
  year        = {2025},
  month       = jan,
  url         = {https://www.acus.gov/sites/default/files/documents/HAI-RegLab-White-Paper-Federal-AI-Leadership-and-Compliance.pdf}
}

@misc{anexries2025exploring,
  author = {Anex-Ries, Quinn},
  title = {Exploring the 2024 Federal {AI} Inventories: Key Improvements, Trends, and Continued Inconsistencies},
  year         = {2025},
  howpublished = {\url{https://cdt.org/insights/exploring-the-2024-federal-ai-inventories-key-improvements-trends-and-continued-inconsistencies/}}
}

@misc{anexries2026oneyear,
  author = {Anex-Ries, Quinn},
  title = {One-Year Retrospective on the Federal Government's Implementation of Updated {AI} Guidance: Accelerating Usage with Incomplete Safeguards},
  year = {2026},
  howpublished = {\url{https://cdt.org/insights/one-year-retrospective-on-the-federal-governments-implementation-of-updated-ai-guidance-accelerating-usage-with-incomplete-safeguards/}}
}

@article{lurie2026triangulating,
  title={Triangulating Across US Federal AI Transparency Regimes},
  author={Lurie, Emma and Fauser, Emma and He, Qing and Metaxa, Dana{\'e} and Friedler, Sorelle A},
  journal={arXiv preprint arXiv:2607.29540},
  year={2026}
}

@article{bryson2017and,
  title={Of, for, and by the people: the legal lacuna of synthetic persons},
  author={Bryson, Joanna J and Diamantis, Mihailis E and Grant, Thomas D},
  journal={Artificial Intelligence and Law},
  volume={25},
  number={3},
  pages={273--291},
  year={2017},
  publisher={Springer}
}

@inproceedings{alanoca2025comparing,
  title={Comparing apples to oranges: A taxonomy for navigating the global landscape of AI regulation},
  author={Alanoca, Sacha and Gur-Arieh, Shira and Zick, Tom and Klyman, Kevin},
  booktitle={Proceedings of the 2025 ACM Conference on Fairness, Accountability, and Transparency},
  pages={914--937},
  year={2025}
}

@techreport{baslandze2026artificial,
  title={Artificial intelligence, productivity, and the workforce: Evidence from corporate executives},
  author={Baslandze, Salom{\'e} and Edwards, Zachary and Graham, John and McClure, Ty and Meyer, Brent H and Sparks, Michael and Waddell, Sonya R and Weitz, Daniel},
  year={2026},
  institution={National Bureau of Economic Research}
}

@misc{gc2026nap,
  title={National Action Plan on Open Government Commitment},
  author={Government of Canada},
  howpublished={\url{https://search.open.canada.ca/nap6//record/2026-06,C03.3.4}},
  year={2026},
  note={[Accessed: Nov 2, 2026]}
}
%TC:endignore

\clearpage
\onecolumn
\appendix
\section{Data Processing Details} % new parent section
\label{app:processing}

\subsection{Fields in Each Release}
\label{app:fields}

Table~\ref{tab:fields-by-year} lists the fields in each public release, which grew from 10 in 2023 to 62 in 2024 and fell to 36 in 2025. Fields asking about the same concept share a row, numbers give each field's position in its file, and a dash means the release did not ask about the concept. The 2023 and 2025 files use short column names, whereas the 2024 file uses the full question text. The last column gives the kind of change described in the Methods. Of the 26 concepts that appear in more than one release, 8 appear in all three, 2 only in 2023 and 2025, and 16 only in 2024 and 2025.

{\scriptsize
\providecommand{\fieldline}[2]{\hangindent=1.3em\hangafter=1\noindent\makebox[1.3em][l]{#1}#2}
\setlength{\LTcapwidth}{\linewidth}
\setlength{\tabcolsep}{4pt}
\begin{longtable}{@{}>{\raggedright\arraybackslash}p{0.14\linewidth} >{\raggedright\arraybackslash}p{0.42\linewidth} >{\raggedright\arraybackslash}p{0.21\linewidth} >{\raggedright\arraybackslash}p{0.16\linewidth}@{}}
  \caption{Fields in each public release of the US Federal AI Use Case Inventory.}
  \label{tab:fields-by-year}\\
  \toprule
  \textbf{2023} (10 fields) & \textbf{2024} (62 fields) & \textbf{2025} (36 fields) & \textbf{Change} \\
  \midrule
  \endfirsthead
  \multicolumn{4}{@{}l}{\emph{Table~\ref{tab:fields-by-year} continued}}\\
  \toprule
  \textbf{2023} (10 fields) & \textbf{2024} (62 fields) & \textbf{2025} (36 fields) & \textbf{Change} \\
  \midrule
  \endhead
  \bottomrule
  \endlastfoot
  \addlinespace[2pt]
  \multicolumn{4}{@{}l@{}}{\parbox{\linewidth}{\raggedright\emph{Identification and organization}}}\\*[1pt]
  \fieldline{1}{\texttt{Use\_\allowbreak{}Case\_\allowbreak{}ID}} & \textendash & \fieldline{3}{\texttt{id}} & Not in the 2024 public file \\
  \fieldline{5}{\texttt{Title}} & \fieldline{1}{Use Case Name} & \fieldline{4}{\texttt{use\_\allowbreak{}case\_\allowbreak{}name}} & Renamed \\
  \fieldline{10}{\texttt{Department}} & \fieldline{2}{Agency} & \fieldline{2}{\texttt{agency\_\allowbreak{}name}} & Renamed \\
  \fieldline{2}{\texttt{Department\_\allowbreak{}Code}} & \fieldline{3}{Agency Abbreviation} & \fieldline{1}{\texttt{agency}} & Renamed \\
  \fieldline{3}{\texttt{Agency}} & \fieldline{4}{Bureau} & \fieldline{5}{\texttt{agency\_\allowbreak{}bureau}} & Renamed \\
  \fieldline{4}{\texttt{Office}} & \textendash & \textendash & Removed \\
  \textendash & \textendash & \fieldline{6}{\texttt{contact\_\allowbreak{}email}} & Introduced \\
  \textendash & \textendash & \fieldline{7}{\texttt{is\_\allowbreak{}withheld}} & Introduced \\
  \addlinespace[2pt]
  \multicolumn{4}{@{}l@{}}{\parbox{\linewidth}{\raggedright\emph{Purpose and design}}}\\*[1pt]
  \fieldline{6}{\texttt{Summary}} & \fieldline{8}{What is the intended purpose and expected benefits of the AI?} & \fieldline{13}{\texttt{problem\_\allowbreak{}solved}}\par \fieldline{14}{\texttt{benefits}} & Split (2024 and 2025) \\
  \fieldline{6}{\texttt{Summary}} & \fieldline{9}{Describe the AI system's outputs.} & \fieldline{15}{\texttt{system\_\allowbreak{}outputs}} & Split (2024); renamed (2025) \\
  \textendash & \fieldline{5}{Use Case Topic Area}\par \fieldline{6}{Other (Use Case Topic Area)} & \fieldline{11}{\texttt{topic\_\allowbreak{}area}} & Introduced (2024); options changed and ``Other'' text dropped (2025) \\
  \fieldline{8}{\texttt{Techniques}} & \textendash & \fieldline{12}{\texttt{classification}} & Removed (2024); added as a fixed classification (2025) \\
  \textendash & \fieldline{7}{Is the AI use case found in the below list of general commercial AI products and services?} & \textendash & Moved to the separate commercial-product inventory (2025) \\
  \addlinespace[2pt]
  \multicolumn{4}{@{}l@{}}{\parbox{\linewidth}{\raggedright\emph{Lifecycle and classification}}}\\*[1pt]
  \fieldline{7}{\texttt{Development\_\allowbreak{}Stage}} & \fieldline{10}{Stage of Development} & \fieldline{8}{\texttt{development\_\allowbreak{}stage}} & Options changed \\
  \textendash & \fieldline{12}{Date Initiated}\par \fieldline{13}{Date when Acquisition and/or Development began}\par \fieldline{14}{Date Implemented}\par \fieldline{15}{Date Retired} & \textendash & Removed \\
  \textendash & \textendash & \fieldline{16}{\texttt{operational\_\allowbreak{}date}} & Introduced \\
  \textendash & \fieldline{11}{Is the AI use case rights-impacting, safety-impacting, both, or neither?} & \fieldline{9}{\texttt{is\_\allowbreak{}high\_\allowbreak{}impact}} & Merged into one high-impact category \\
  \textendash & \textendash & \fieldline{10}{\texttt{HI\_\allowbreak{}justification}} & Introduced \\
  \addlinespace[2pt]
  \multicolumn{4}{@{}l@{}}{\parbox{\linewidth}{\raggedright\emph{Procurement}}}\\*[1pt]
  \textendash & \fieldline{16}{Was the AI system involved in this use case developed (or is it to be developed) under contract(s) or in-house?} & \fieldline{17}{\texttt{contracting\_\allowbreak{}usage}} & Options changed: ``purchased from a vendor'' \\
  \textendash & \fieldline{17}{Provide the Procurement Instrument Identifier(s) (PIID) of the contract(s) used.} & \fieldline{18}{\texttt{vendor\_\allowbreak{}name}} & Reframed: contract identifier became vendor name; not compared \\
  \addlinespace[2pt]
  \multicolumn{4}{@{}l@{}}{\parbox{\linewidth}{\raggedright\emph{Public-facing services}}}\\*[1pt]
  \textendash & \fieldline{18}{Is this AI use case supporting a High-Impact Service Provider (HISP) public-facing service?}\par \fieldline{19}{Which HISP is the AI use case supporting?}\par \fieldline{20}{Which public-facing service is the AI use case supporting?} & \textendash & Removed \\
  \textendash & \fieldline{21}{Does this AI use case disseminate information to the public?}\par \fieldline{22}{How is the agency ensuring compliance with Information Quality Act guidelines, if applicable?} & \textendash & Removed \\
  \addlinespace[2pt]
  \multicolumn{4}{@{}l@{}}{\parbox{\linewidth}{\raggedright\emph{Data and privacy}}}\\*[1pt]
  \textendash & \fieldline{23}{Does this AI use case involve personally identifiable information (PII) that is maintained by the agency?} & \fieldline{23}{\texttt{has\_\allowbreak{}pii}} & Renamed \\
  \textendash & \fieldline{24}{Has the Senior Agency Official for Privacy (SAOP) assessed the privacy risks associated with this AI use case?} & \fieldline{24}{\texttt{pia\_\allowbreak{}url}} & Replaced: privacy review became a PIA link; not compared \\
  \textendash & \fieldline{25}{Do you have access to an enterprise data catalog or agency-wide data repository that enables you to identify whether or not the necessary datasets exist and are ready to develop your use case?}\par \fieldline{26}{If Other, please explain.} & \fieldline{22}{\texttt{link\_\allowbreak{}to\_\allowbreak{}data}} & Replaced: catalog access became a catalog link; not compared \\
  \textendash & \fieldline{27}{Describe any agency-owned data used to train, fine-tune, and/or evaluate performance of the model(s) used in this use case.} & \fieldline{21}{\texttt{data\_\allowbreak{}description}} & Reworded: any data, not only agency-owned \\
  \textendash & \fieldline{28}{Is there available documentation for the model training and evaluation data that demonstrates the degree to which it is appropriate to be used in analysis or for making predictions?} & \textendash & Removed \\
  \textendash & \fieldline{29}{Which, if any, demographic variables does the AI use case explicitly use as model features?}\par \fieldline{30}{If Other, please explain.} & \fieldline{25}{\texttt{demographic\_\allowbreak{}features}} & Options changed; ``Other'' text dropped \\
  \addlinespace[2pt]
  \multicolumn{4}{@{}l@{}}{\parbox{\linewidth}{\raggedright\emph{Code and authorization}}}\\*[1pt]
  \textendash & \fieldline{31}{Does this project include custom-developed code?} & \fieldline{26}{\texttt{has\_\allowbreak{}custom\_\allowbreak{}code}} & Renamed \\
  \textendash & \fieldline{32}{Does the agency have access to the code associated with the AI use case?} & \textendash & Removed \\
  \fieldline{9}{\texttt{Source\_\allowbreak{}Code}} & \fieldline{33}{If the code is open-source, provide the link for the publicly available source code.} & \fieldline{27}{\texttt{code\_\allowbreak{}url}} & Renamed \\
  \textendash & \fieldline{34}{Does this AI use case have an associated Authority to Operate (ATO) for an AI system?} & \fieldline{19}{\texttt{have\_\allowbreak{}ato}} & Renamed \\
  \textendash & \fieldline{35}{System Name} & \fieldline{20}{\texttt{system\_\allowbreak{}name\_\allowbreak{}ato}} & Renamed \\
  \addlinespace[2pt]
  \multicolumn{4}{@{}l@{}}{\parbox{\linewidth}{\raggedright\emph{Computing resources and internal sharing}}}\\*[1pt]
  \textendash & \fieldline{36}{How long have you waited for the necessary developer tools to implement the AI use case?}\par \fieldline{37}{For this AI use case, is the required IT infrastructure provisioned via a centralized intake form or process inside the agency?}\par \fieldline{38}{If Other, please explain.}\par \fieldline{39}{Do you have a process in place to request access to computing resources for model training and development of the AI involved in this use case?}\par \fieldline{40}{If Other, please explain.}\par \fieldline{41}{Has communication regarding the provisioning of your requested resources been timely?}\par \fieldline{42}{If Other, please explain.}\par \fieldline{43}{How are existing data science tools, libraries, data products, and internally-developed AI infrastructure being re-used for the current AI use case?}\par \fieldline{44}{Has information regarding the AI use case, including performance metrics and intended use of the model, been made available for review and feedback within the agency?} & \textendash & Removed \\
  \addlinespace[2pt]
  \multicolumn{4}{@{}l@{}}{\parbox{\linewidth}{\raggedright\emph{Risk-management practices (required only of rights- or safety-impacting uses in 2024 and of deployed high-impact uses in 2025)}}}\\*[1pt]
  \textendash & \fieldline{45}{Has your agency requested an extension to implement the minimum risk management practices for this AI use case?} & \textendash & Removed \\
  \textendash & \fieldline{46}{Has an AI impact assessment been conducted for this AI use case?} & \fieldline{29}{\texttt{hi\_\allowbreak{}assessment\_\allowbreak{}completed}} & Options changed \\
  \textendash & \fieldline{47}{Has the AI use case been tested in operational or real-world environments to understand the performance and impact it may have on affected individuals or communities?} & \fieldline{28}{\texttt{hi\_\allowbreak{}testing\_\allowbreak{}conducted}} & Reworded: pre-deployment testing \\
  \textendash & \fieldline{48}{What are the key risks from using the AI for this particular use case and how were they identified?} & \fieldline{30}{\texttt{hi\_\allowbreak{}potential\_\allowbreak{}impacts}} & Reworded: potential impacts \\
  \textendash & \fieldline{49}{Has an independent evaluation of the AI use case been conducted?} & \fieldline{31}{\texttt{hi\_\allowbreak{}independent\_\allowbreak{}review}} & Reworded: ``review'' \\
  \textendash & \fieldline{50}{Is there a process to monitor performance of the AI system's functionality and changes to its impact on rights or safety as part of the post-deployment plan for the AI use case?} & \fieldline{32}{\texttt{hi\_\allowbreak{}ongoing\_\allowbreak{}monitoring}} & Options changed \\
  \textendash & \fieldline{51}{For this particular use case, can the AI carry out a decision or action without direct human involvement that could result in a significant impact on rights or safety?}\par \fieldline{52}{If Other, please explain.} & \textendash & Removed \\
  \textendash & \textendash & \fieldline{33}{\texttt{hi\_\allowbreak{}training\_\allowbreak{}established}} & Introduced \\
  \textendash & \textendash & \fieldline{34}{\texttt{hi\_\allowbreak{}failsafe\_\allowbreak{}presence}} & Introduced \\
  \textendash & \fieldline{53}{How is the agency providing reasonable and timely notice regarding the use of AI when people interact with an AI-enabled service as a result of this AI use case?}\par \fieldline{54}{If Other, please explain.} & \textendash & Removed \\
  \textendash & \fieldline{55}{Is the AI used to significantly influence or inform decisions or actions that could have an adverse or negative impact on specific individuals or groups?} & \textendash & Removed \\
  \textendash & \fieldline{56}{What steps has the agency taken to detect and mitigate significant disparities in the model's performance across demographic groups for this AI use case?} & \textendash & Removed \\
  \textendash & \fieldline{57}{What steps has the agency taken to consult and incorporate feedback from groups affected by this AI use case?}\par \fieldline{58}{If Other, please explain.} & \fieldline{36}{\texttt{hi\_\allowbreak{}public\_\allowbreak{}consultation}} & Reworded: end users and the public \\
  \textendash & \fieldline{59}{Is there an established fallback and escalation process for this AI use case in the event that an impacted individual or group would like to appeal or contest the AI system's outcome?}\par \fieldline{60}{If No, please explain.} & \fieldline{35}{\texttt{hi\_\allowbreak{}appeal\_\allowbreak{}process}} & Reworded: appeal process; explanation dropped \\
  \textendash & \fieldline{61}{Where practicable and consistent with applicable law and governmentwide policy, is there an established mechanism for individuals to opt-out from the AI functionality in favor of a human alternative?}\par \fieldline{62}{If Other, please explain.} & \textendash & Removed \\
\end{longtable}
}

\begin{table}[!h]
  \caption{How the inventory's questions changed across releases, with an example of each.}
  \label{tab:question-changes}
  \scriptsize
  \setlength{\tabcolsep}{4pt}
  \renewcommand{\arraystretch}{1.15}
  \begin{tabular}{@{}>{\raggedright\arraybackslash}p{0.12\linewidth} >{\raggedright\arraybackslash}p{0.3\linewidth} >{\raggedright\arraybackslash}p{0.52\linewidth}@{}}
    \toprule
    Change & What happened & Example \\
    \midrule
    Introduced & A later release asked a question with no earlier counterpart. & 2025 asked whether operators of high-impact (HI) uses had been trained. \\
    \addlinespace
    Removed & An earlier question had no later counterpart. & 2024 asked whether a use could act without human involvement; 2025 did not. \\
    \addlinespace
    Renamed & The same question appeared under a different field name. & 2024 \texttt{Bureau} became 2025 \texttt{agency\_bureau}. \\
    \addlinespace
    Split & One question became two or more. & The 2023 summary became separate 2024 questions on purpose and benefits and on outputs; 2025 then split purpose and benefits into the problem addressed and the expected benefits. \\
    \addlinespace
    Merged & Two or more questions or categories became one. & 2024 distinguished rights- and safety-impacting categories; 2025 combined them into a single HI category. \\
    \addlinespace
    Moved & A question left the use-case file for a separate one. & 2024 asked whether each use was a common commercial AI product; 2025 reported such products in the separate COTS inventory. \\
    \bottomrule
  \end{tabular}
\end{table}

\clearpage

\subsection{Labeling Whether an Entry Had to Answer}
\label{app:requirement_check}

We read each year's data dictionary alongside OMB guidance to find which entries each question applied to. Most conditions depended on an entry's lifecycle stage, its impact classification, or whether it relied solely on a listed commercial product. We labeled each question for each entry as \emph{required}, \emph{not required}, \emph{optional}, or \emph{unspecified}. A label describes the requirement, not the answer: an entry labeled \emph{required} could still leave the question blank. Table~\ref{tab:requirement-examples} shows how we applied these rules to one VA entry and to two 2024 entries whose label we could not determine.

\begin{table}[!ht]
  \caption{Worked examples of how we decided whether an entry had to answer a question.}
  \label{tab:requirement-examples}
  \scriptsize
  \setlength{\tabcolsep}{4pt}
  \renewcommand{\arraystretch}{1.15}
  \begin{tabular}{@{}>{\raggedright\arraybackslash}p{0.22\linewidth} >{\raggedright\arraybackslash}p{0.2\linewidth} >{\raggedright\arraybackslash}p{0.4\linewidth} >{\raggedright\arraybackslash}p{0.12\linewidth}@{}}
    \toprule
    Entry & Question & Why & Label \\
    \midrule
    VA, GE Logiq E10 (VA-24-204), 2025 & How the use was procured & Required of pilot and deployed uses; this use was deployed (it answered ``Vendor Purchased''). & Required$^{a}$ \\
    \addlinespace
    Same entry & Whether the use involves PII & Required of pilot and deployed uses; the entry left it blank. & Required$^{a}$ \\
    \addlinespace
    Same entry & Whether an impact assessment was completed & Required only of deployed high-impact uses; this use was classified as not high-impact. & Not required$^{b}$ \\
    \addlinespace
    Same entry & Link to public source code & The data dictionary required it of no entry. & Optional$^{c}$ \\
    \addlinespace
    NASA, Enhanced AutoNav for Perseverance Rover on Mars, 2024 & Whether the use involves PII & Required of uses in implementation or operation; the entry's stage, ``In production,'' was not one of the inventory's five stages. & Unspecified$^{d}$ \\
    \addlinespace
    Commerce, GitHub Copilot for Code Modernization, 2024 & Whether the use involves PII & Required of uses in operation unless built solely from a listed commercial product; the entry left the commercial-product question blank, so we could not tell whether it was exempt. & Unspecified$^{d}$ \\
    \bottomrule
  \end{tabular}

  \vspace{2pt}
  \begin{minipage}{\linewidth}\scriptsize
  $^{a}$\emph{Required}: that year's rules required the entry to answer, whether or not it did;\\
  $^{b}$\emph{not required}: the entry's own answers placed it outside a conditional requirement;\\
  $^{c}$\emph{optional}: the data dictionary required the question of no entry;\\
  $^{d}$\emph{unspecified}: a missing or nonstandard answer kept us from deciding, such as a stage that did not match the year's options.
  \end{minipage}
\end{table}

\subsection{Response Classification}
\label{app:response-classification}

After normalizing whitespace, case, and typographic dashes, the classifier applied field-specific overrides and listed response tokens. \emph{Blank} covered empty values and the placeholders \texttt{[]}, \texttt{\{\}}, \texttt{null}, and \texttt{nan}; \emph{Not applicable}, the tokens \texttt{na}, \texttt{n/a}, \texttt{n.a.}, \texttt{not applicable}, and \texttt{not-applicable}; \emph{Unknown}, ``unknown,'' ``undetermined,'' ``to be determined,'' ``tbd,'' ``not available,'' and the exact answer \texttt{Unknown, see "Comments" field}; and \emph{Withheld}, standalone withholding and redaction notices such as ``withheld,'' ``not publicly releasable,'' ``REDACTED,'' and ``Redacted for cybersecurity purposes.'' \emph{No} covered ``no,'' ``false,'' ``none,'' ``none of the above,'' and responses beginning with ``no~'' or ``no~-.'' All other non-blank responses were \emph{Present}, so an answer beginning with ``N/A'' and continuing with an explanation remained \emph{Present} unless it matched another rule. The classifier did not judge the content of any response.

Of the 12 columns that recurred between 2024 and 2025, Table~\ref{tab:disclosure} reports 10. No matched use had to give a source-code link in both years, and the 2024 contract identifiers and the 2025 vendor names asked for different information. In Table~\ref{tab:disclosure}, \emph{Other} counts changes that involved an \emph{Unknown}, \emph{Not applicable}, \emph{Withheld}, or \emph{Invalid} answer; we counted these as neither gains nor losses. Before counting distinct justifications for PHIN classifications, we standardized characters and spacing; the 109 justifications contained 69 distinct texts.

\subsection{Record Linkage}
\label{app:methods-supplement}

We compared entries only within the same agency, using canonical agency codes and treating DOS and STATE as one agency. Comparison strings were whitespace-collapsed, lowercased, and reduced to ASCII letters and digits; the original values were kept for review. For titles, we also removed standalone ``AI'' and ``artificial intelligence,'' inventory annotations, and leading ``Renamed:'' markers, and kept both the full title and the names on either side of a ``Previously:'' marker as aliases. For identifiers and bureaus, we also replaced ampersands with ``and'' and removed ``the,'' ``of,'' ``for,'' ``a,'' and ``an.''

Deterministic matching then took, in order and among still-unmatched records, (1) agency--identifier matches; (2) agency--title-alias matches; (3) repeated exact titles told apart by equivalent bureau labels (equal after normalization, one label of at least three characters contained in the other, or one label's token set contained in the other's); and (4) pairs in which one title alias began with the other and the shorter alias had at least 40 characters. Every deterministic match had to be unique in both directions. Because the 2024 file had no identifier column and the 2023 and 2025 identifiers used different formats, no accepted link came from identifiers (Table~\ref{tab:methods-linkage-review}).

For the remaining records, title and bureau scores combined character-sequence similarity ($R$, from Python's \texttt{difflib.SequenceMatcher}), token-set Jaccard overlap ($J$, shared tokens over their union), and token containment ($C$, shared tokens over the smaller set) as $0.45R + 0.35J + 0.20C$, taking the highest-scoring pair of title aliases. Description scores compared the normalized token sets of the 2023 summary, the combined 2024 purpose-and-benefits and output fields, and the combined 2025 problem, benefits, and output fields as $0.60J+0.40C$. The composite score was
\begin{equation}
S = 0.72S_{\mathrm{title}} + 0.20S_{\mathrm{description}} + 0.08S_{\mathrm{bureau}}.
\label{eq:methods-linkage-score}
\end{equation}
Empty comparisons contributed zero, without redistributing their weights. An exact normalized title-alias match raised $S$ to at least 0.98, and an exact nonempty normalized identifier match set it to 1.00. These scores ranked candidates; they are not calibrated match probabilities.

Maximum-weight Hungarian assignment produced one-to-one candidates within each agency. A candidate pair advanced to author review if $S\geq0.48$ and either $S_{\mathrm{title}}\geq0.50$, or both $S_{\mathrm{title}}\geq0.35$ and $S_{\mathrm{description}}\geq0.45$. Only accepted candidates entered the panels alongside deterministic matches (Table~\ref{tab:methods-linkage-review}). Joining the 2023--2024 and 2024--2025 links through the same 2024 entry followed 150 uses across all three releases: 143 agreed with a direct 2023--2025 link, 6 had no direct link, and 1 conflicted with its direct link.

\begin{table}[htbp]
    \centering
    \caption{Deterministic matches by method, candidate-review decisions, and accepted links. We constructed the three panels separately. No deterministic match came from an identifier.}
    \label{tab:methods-linkage-review}
    \small
    \setlength{\tabcolsep}{4pt}
    \begin{tabular}{lrrrrrrrr}
        \toprule
        & \multicolumn{3}{c}{Deterministic} & \multicolumn{4}{c}{Candidate review} & \\
        \cmidrule(lr){2-4}\cmidrule(lr){5-8}
        Panel & Title & \shortstack{Title +\\bureau} & \shortstack{Title\\prefix} & Reviewed & Accepted & Rejected & Uncertain & Total links \\
        \midrule
        2023--2024 & 210 & 0 & 2 & 44 & 42 & 1 & 1 & 254 \\
        2024--2025 & 1,126 & 18 & 18 & 186 & 170 & 16 & 0 & 1,332 \\
        2023--2025 & 187 & 0 & 2 & 43 & 38 & 5 & 0 & 227 \\
        \bottomrule
    \end{tabular}
\end{table}

\subsection{Keyword Search of Benefits} % new
\label{app:benefits-keywords}

For the values that agencies claimed, we searched the 2,821 non-blank 2025 answers about benefits, among entries required to answer, with the case-insensitive patterns in Table~\ref{tab:benefit-keywords}. An answer could match more than one value.

\begin{table}[!ht]
  \caption{Keyword patterns for the values claimed in 2025 benefits. An asterisk matches any word ending.}
  \label{tab:benefit-keywords}
  \scriptsize
  \setlength{\tabcolsep}{4pt}
  \renewcommand{\arraystretch}{1.15}
  \begin{tabular}{@{}>{\raggedright\arraybackslash}p{0.17\linewidth} >{\raggedright\arraybackslash}p{0.64\linewidth} r@{}}
    \toprule
    Value & Patterns & Answers (\%) \\
    \midrule
    Efficiency & efficien*, productiv*, cost(s), cost sav*, time sav*, time-consuming, workload, streamlin*, faster, quicker, accelerat*; save(s) or saving(s) + time, money, hours, staff, or cost(s); reduc* (the) + time, cost(s), manual, workload, effort, burden, labor, or hours; manual + effort, work, review, process*, labor, or task(s) & 1,260 (44.7) \\
    \addlinespace
    Accuracy or consistency & accura*, consisten* (not ``consistent with''), error(s) & 424 (15.0) \\
    \addlinespace
    Access to services & access to, accessib*, customer service, customer experience, service to the public & 160 (5.7) \\
    \addlinespace
    Safety or security & safety, safer, security & 173 (6.1) \\
    \addlinespace
    Equity or fairness & equit* (not ``equity price''), fair* & 13 (0.5) \\
    \bottomrule
  \end{tabular}
\end{table}

\section{Supplementary Results}
\label{app:results}

This section reports the lifecycle cross-tabulation described in the Methods, PII use and the demographic variables entries named (Figure~\ref{fig:app-pii-demographics}), and figures behind counts reported in the Results. Most uses deployed in 2024 remained deployed in 2025 (450 of 584, 77.1\%), whereas 128 of the 543 uses in pre-deployment in 2024 (23.6\%) were retired by 2025 (Table~\ref{tab:results-lifecycle}).

\begin{table}[!ht]
  \caption{Lifecycle stage of the 1,332 matched uses in 2024 (rows) and 2025 (columns).}
  \label{tab:results-lifecycle}
  \centering\small
  \setlength{\tabcolsep}{4pt}
  \begin{tabular}{@{}lrrrrrr@{}}
    \toprule
    & \multicolumn{5}{c}{2025 stage} & \\
    \cmidrule(lr){2-6}
    2024 stage & \shortstack{Pre-\\deployment} & Pilot & Deployed & Retired & \shortstack{Unknown/\\other} & Total \\
    \midrule
    Pre-deployment & 269 & 51 & 75 & 128 & 20 & 543 \\
    Pilot & 21 & 62 & 51 & 29 & 8 & 171 \\
    Deployed & 16 & 5 & 450 & 80 & 33 & 584 \\
    Retired & 0 & 1 & 2 & 21 & 2 & 26 \\
    Unknown/other & 7 & 0 & 1 & 0 & 0 & 8 \\
    \midrule
    Total & 313 & 119 & 579 & 258 & 63 & 1,332 \\
    \bottomrule
  \end{tabular}
  \par\smallskip\begin{minipage}{\linewidth}\footnotesize Stages are grouped into four common stages as described in the Methods. \emph{Unknown/other} covers blank stages and stages that matched none of the year's options.\end{minipage}
\end{table}

\begin{figure}[!ht]
  \centering
  \includegraphics[width=\linewidth]{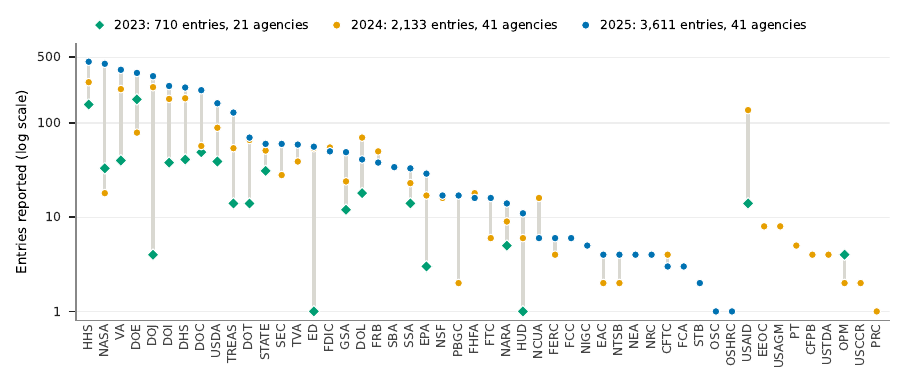}
  \caption{Number of AI use cases reported by different departments and agencies in different years.}
  \label{fig:agencies}
\end{figure}

\begin{figure}[!ht]
    \centering
    \includegraphics[width=.8\linewidth]{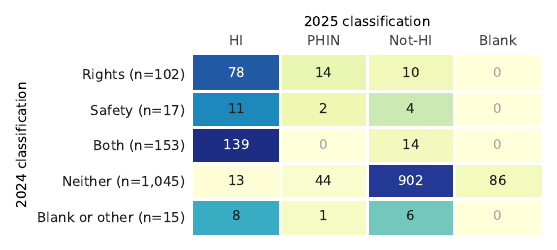}
    \caption{Impact classification transition between 2024 and 2025 among matched uses.}
    \label{fig:classification_transition}
\end{figure}

% Moved from the old Figure 3 (c) and (d). The subfigure labels are kept, so the Results' references to
% fig:techniques-pii and fig:techniques-demographics now point here without further edits.
\begin{figure}[htbp]
  \centering
  \includegraphics[width=.6\linewidth]{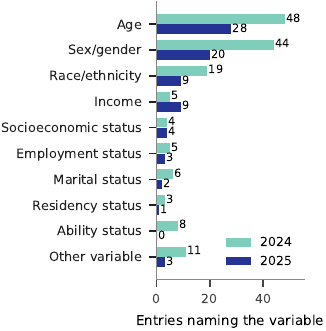}
    \caption{Demographic variables named as model features.}
    \label{fig:demographics}
\end{figure}

\end{document}